\documentclass[runningheads]{llncs}

\usepackage{eccv}

\usepackage{eccvabbrv}

\usepackage{graphicx}
\usepackage{booktabs}

\usepackage[accsupp]{axessibility}  

\usepackage[pagebackref]{hyperref}

\usepackage{orcidlink}

\usepackage{booktabs}   
\usepackage{siunitx}
\usepackage{multirow}   
\usepackage{graphicx}   
\usepackage{amsmath}    
\usepackage{caption}    

\usepackage{float}  

\usepackage{amssymb}
\usepackage{booktabs}

\usepackage{algorithm}
\usepackage{algorithmic}

\usepackage{listings}
\usepackage{tabularx}
\usepackage{pifont}

\usepackage{longtable}
\usepackage{booktabs}
\usepackage{pifont}
\usepackage{xcolor}
\usepackage{placeins}
\usepackage{tcolorbox}
\tcbuselibrary{breakable}
\newcommand{\cmark}{\textcolor{green!60!black}{\ding{51}}}
\newcommand{\xmark}{\textcolor{red}{\ding{55}}}
\usepackage[table]{xcolor}

\newcommand{\figsep}[1]{
    \vspace{10pt}
    
    \noindent\textbf{Example #1}
    \hrule
    \vspace{12pt}
}

\usepackage{array}
\definecolor{lightgrayrow}{gray}{0.94}
\newcommand{\uparrowgreen}{\textcolor{green!50!black}{$\uparrow$}}
\newcommand{\downarrowred}{\textcolor{red!65!black}{$\downarrow$}}
\usepackage{tikz}
\usetikzlibrary{positioning,arrows.meta,fit,matrix} 
\begin{document}

\title{RealCQA-V2: A Diagnostic Benchmark for Structured Visual Entailment over Scientific Charts} 

\titlerunning{RealCQA-V2:VPP}




\author{
Saleem Ahmed \and
Srirangaraj Setlur \and
Venu Govindaraju
}

\authorrunning{S. Ahmed et al.}

\institute{
University at Buffalo, Buffalo, NY, USA\\
\email{sahmed9@buffalo.edu}
}

\maketitle

\begin{abstract}
Multimodal reasoning models often produce fluent answers supported by seemingly coherent rationales. Existing benchmarks evaluate only final-answer correctness. They do not support atomic visual entailment verification of intermediate steps, especially visual compositional logic. This limitation is especially acute in scientific chart understanding, where answers depend on deterministically grounded visual semantics such as axes, legends, and quantitative relations.
We introduce RealCQA-V2, a large-scale benchmark that reformulates chart question answering as \emph{Visual Premise Proving (VPP)}: a structured logical entailment task over chart-grounded visual predicates. Each question is deconstructed into manually curated, atomic premises grounded in chart elements (axes, legends, marks, and quantitative relations), yielding executable reasoning chains rather than free-form textual rationales. These premises form compositional reasoning chains, enabling verification at the level of individual visual statements and complete reasoning sequences.
We introduce chain-level metrics that measure both full logical validity (\textit{Acc\textsubscript{VPP}}) and partial reasoning progress within failed chains (\textit{DCP}), extending beyond traditional VQA accuracy. Baseline evaluations across representative LVLMs reveal a consistent local–global reasoning gap: models often verify many individual premises correctly while failing to preserve coherence across the full chain.
RealCQA-V2 establishes a reproducible benchmark for structured visual entailment over real scientific charts and enables rigorous diagnosis of multimodal reasoning beyond answer-only evaluation.
\end{abstract}

\section{Introduction}

\begin{figure}[t]
\centering
\includegraphics[width=\linewidth]{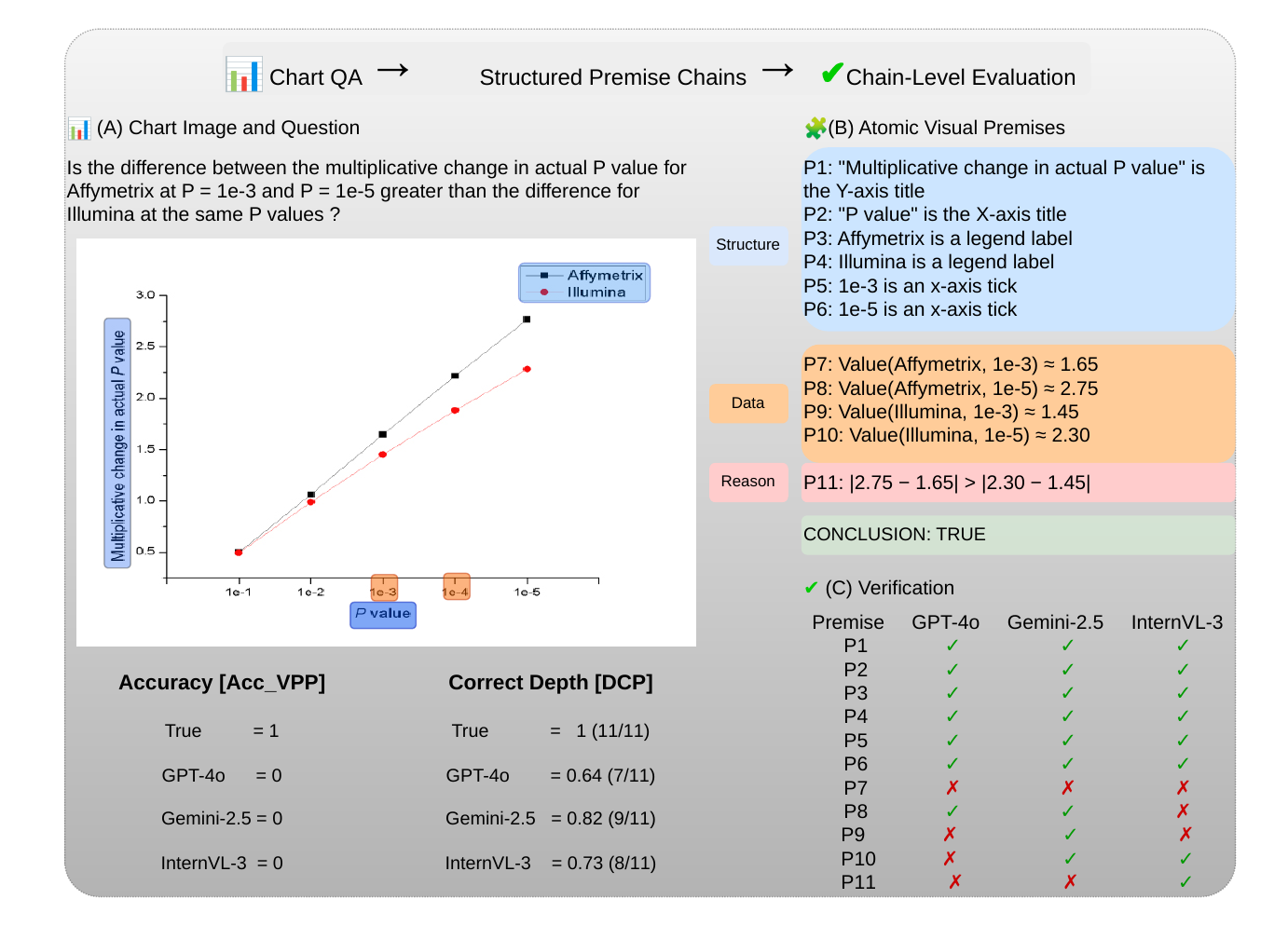}
\caption{
\textbf{Visual Premise Proving (VPP)}. A chart question is decomposed into atomic chart-grounded premises (e.g., structural, data, and reasoning steps), and models verify the resulting premise chain via binary entailment judgments.
}
\label{fig:vpp_overview}
\end{figure}

Chain-of-Thought (CoT) prompting has made intermediate reasoning explicit in large language models (LLMs), improving performance on complex tasks while offering an interpretable decomposition of decisions~\cite{wei2022chain,kojima2022large,wang2022self}. As vision--language models move toward general-purpose agents, similar stepwise rationales are increasingly produced for multimodal inputs. Yet existing benchmarks largely remain answer-centric: they rarely test whether intermediate steps are \emph{visually entailed} by the underlying image, and thus cannot distinguish faithful reasoning from plausible narration.

This gap is now a practical limitation. Large vision--language models (LVLMs) can generate fluent, image-referential explanations, but these rationales are seldom verifiable at the level of individual inferences. Prior work shows that LVLMs may reach correct answers through hallucinated or incoherent intermediate logic, including steps that are unsupported by the visual evidence~\cite{zhang2023multimodal}. Without structured verification, explanation quality and reasoning faithfulness become difficult to measure, compare, or improve.

A core bottleneck is the lack of benchmarks that operationalize multimodal reasoning as a \emph{structured entailment problem}. Existing datasets emphasize end-task accuracy and provide limited supervision for intermediate inference, conflating correct answers with correct reasoning. They do not permit atomic, visually grounded verification of reasoning steps, nor do they measure compositional depth and consistency across multi-step chains. As a result, progress toward verifiable multimodal reasoning is difficult to evaluate reliably.

Scientific charts provide a uniquely suitable testbed for this problem. Charts expose explicit semantic variables---axes, legends, scales, and marks---whose values can be grounded deterministically, while answering chart questions requires compositional operations such as value retrieval, comparison, aggregation, and trend analysis. This combination of structured semantics and multi-step composition makes charts an ideal domain for studying \emph{verifiable visual entailment}.

To address this gap, we introduce \textbf{Visual Premise Proving (VPP)}, which formulates chart reasoning as logical entailment evaluation over explicit, chart-grounded visual predicates. To instantiate VPP, we construct \textbf{RealCQA-V2}, a large-scale benchmark in which each chart question is distilled into a sequence of manually curated atomic premises grounded in chart components and quantitative relations. These premises form structured reasoning chains that can be verified step-by-step, enabling evaluation of not only \emph{what} a model answers, but \emph{whether the intermediate reasoning is entailed by the chart}.

RealCQA-V2 contains 5M premise--conclusion pairs spanning 1.7M questions over 28K real-world charts curated from PubMed Central, covering diverse chart types, reasoning depths, and question formats. We introduce chain-level metrics that measure complete logical validity ($Acc_{\textsubscript{VPP}}$) and partial reasoning progress within failed chains (DCP). Together, these metrics expose a measurable local--global reasoning gap: high premise-level correctness can coexist with low full-chain consistency, revealing failure modes that are invisible under answer-only evaluation.

\noindent\textbf{Our contributions are threefold:}
\begin{itemize}
    \item \textbf{A large-scale benchmark for atomic visual entailment.} 
    We introduce RealCQA-V2, a dataset that reformulates chart question answering as Visual Premise Proving (VPP) by distilling each question into manually curated, chart-grounded atomic premises that support structured entailment evaluation.

    \item \textbf{A reproducible chain-level evaluation protocol.} 
    We define standardized metrics ($Acc_{\textsubscript{VPP}}$ and DCP) that measure full-chain validity and partial reasoning progress, enabling fine-grained, reproducible verification beyond final-answer accuracy.

    \item \textbf{A controlled testbed that reveals a measurable local--global reasoning gap in LVLMs.} 
    Built from 28K real-world scientific charts with 5M premise--conclusion pairs, RealCQA-V2 provides a scalable framework for diagnosing how local premise verification succeeds or fails to compose into globally consistent reasoning.
\end{itemize}
\section{Background}
\label{sec:background}

We review prior work on Chain-of-Thought (CoT) reasoning, multimodal reasoning benchmarks, and chart-based visual reasoning.

\subsection{CoT Reasoning in Language Models}
\label{sec:cot-text}

Chain-of-Thought (CoT) prompting has become a central technique for eliciting stepwise reasoning in large language models (LLMs), shifting the focus from direct answer prediction toward structured intermediate deduction~\cite{wei2022chain}. CoT improves performance on tasks involving arithmetic, symbolic manipulation, and commonsense inference~\cite{kojima2022large,wang2022self}.

Subsequent work explored ways to improve the reliability and efficiency of CoT, including distillation~\cite{magister2022teaching}, exemplar selection through retrieval or clustering~\cite{rubin2022learning,zhang2022automatic}, and structured prompting strategies such as least-to-most reasoning~\cite{zhou2022least}, program-of-thought reasoning~\cite{chen2022program}, and self-consistency decoding~\cite{wang2022self}. While these advances strengthen reasoning in text-only settings, extending CoT to multimodal inputs introduces an additional requirement: intermediate steps must be grounded in observable visual evidence rather than only text.

\subsection{Multimodal CoT: Benchmarks and Gaps}
\label{sec:cot-multimodal}

Several benchmarks extend CoT-style reasoning to multimodal tasks by pairing visual inputs with explanatory rationales. \textbf{ScienceQA}~\cite{lu2022learn} and \textbf{A-OKVQA}~\cite{schwenk2022okvqa}, for example, provide image-based questions with textual explanations and show that rationale supervision can improve answer accuracy. However, these explanations are typically treated as free-form justifications rather than explicit reasoning structures aligned to visual semantics.

Other benchmarks move closer to entailment-style evaluation. \textbf{PMR}~\cite{dong2021premise} formulates image--text reasoning as premise--hypothesis entailment, while \textbf{VCR}~\cite{zellers2019recognition} links rationales to specific image regions to improve traceability between language and visual evidence. These datasets provide stronger grounding than free-form explanation benchmarks, but their reasoning structures remain relatively coarse and are difficult to verify at the level of individual inference steps.

Recent modeling work applies LVLMs such as Flamingo and GPT-4V to generate multimodal CoTs directly, or fine-tunes open-source architectures such as T5 in the \textbf{Multimodal-CoT} framework~\cite{zhang2023multimodal}. Despite promising performance, these systems continue to exhibit hallucinated or logically inconsistent intermediate steps.

Across these datasets and modeling efforts, three limitations persist. First, rationales often lack \textit{atomic structure}: they are not decomposed into minimal reasoning units that can be independently verified. Second, intermediate steps may violate \textit{visual entailment}, meaning that reasoning statements are not fully grounded in observable evidence~\cite{xie2019visual}. Third, reasoning rarely follows a \textit{canonical compositional structure} that chains atomic facts into coherent proofs~\cite{johnson2017clevr}. Together, these limitations make it difficult to evaluate whether multimodal reasoning is both locally grounded and globally coherent.

\subsection{Chart-Based Visual Reasoning}
\label{sec:chart-reasoning}

Charts are structured visual artifacts that encode quantitative relationships through axes, legends, scales, and graphical marks. Their explicit semantic organization makes them a natural testbed for studying grounded, multi-step visual reasoning. Several datasets have explored chart understanding from different perspectives. Early work such as \textbf{FigureSeer}~\cite{Siegel2016FigureSeerPR} and \textbf{ChartInfo}~\cite{davila2022icpr} focuses on dense chart parsing, including component detection, text extraction, and structural understanding.

Chart-based question answering datasets such as \textbf{ChartQA}~\cite{masry2022chartqa} and \textbf{CharXiv}~\cite{wang2024charxiv} evaluate reasoning over chart images using natural-language queries. However, many existing chart benchmarks are either synthetic or only weakly annotated. Synthetic datasets such as \textbf{FigureQA}~\cite{kahou2018figureqa}, \textbf{DVQA}, and \textbf{LeafQA} scale easily by rendering charts from tabular data, but they lack the visual complexity, annotation fidelity, and semantic richness of real scientific figures.

Real-world scientific charts are substantially more challenging: they often contain dense layouts, overlapping plot elements, specialized notation, and domain-specific semantics. \textbf{RealCQA}~\cite{ahmed2023realcqa}, built on the manually annotated \textbf{ChartInfo} dataset~\cite{davila2022icpr}, provides question-answer templates grounded in chart components extracted from PubMed Central figures. These real world charts come with dense annotations of axes, tick values, legend entries, and data series, thus supporting precise grounding between visual elements and semantic variables.

These properties make RealCQA uniquely suitable for constructing a benchmark in which reasoning statements are semantically meaningful and directly verifiable against chart annotations. The chart components provide a closed set of variables, such as axis ticks, legend labels, and data values. While template-based question generation induces a corresponding set of predicates describing relations among those variables. This combination enables the construction of formally verifiable reasoning sequences grounded in real-world chart semantics, which forms the basis of our Visual Premise Proving framework.

\section{Visual Premise Proving (VPP)}
\label{sec:vpp}

We introduce \textbf{Visual Premise Proving (VPP)}, a benchmark task that evaluates multimodal reasoning over charts as logical verification of intermediate visual statements. Traditional chart question answering benchmarks assess only final-answer correctness, allowing models to appear successful even when intermediate reasoning is unsupported by the visual evidence. VPP instead asks whether each intermediate reasoning statement is entailed by the chart.

Given a chart $C$ and a question $Q$, the benchmark provides an ordered premise sequence
\[
\mathcal{P} = (p_1, p_2, \ldots, p_n),
\]
where each premise is an atomic reasoning statement grounded in chart structure, numerical values, or derived quantitative relations. In this sense, VPP formulates chart reasoning as structured entailment evaluation over chart-grounded predicates. A reasoning chain is valid if every premise is supported by the chart:
\[
C \models p_i \quad \forall p_i \in \mathcal{P}.
\]
This formulation exposes reasoning failures that remain hidden under answer-only evaluation. Figure~\ref{fig:vpp_overview} illustrates how conventional chart QA is transformed into premise-level verification, enabling fine-grained reasoning diagnosis.

VPP separates three capabilities that are typically conflated in existing benchmarks: perception of chart structure, retrieval of numerical values, and logical composition of relationships among those values. By evaluating these components separately yet within a common chain structure, the benchmark enables diagnostic analysis of multimodal reasoning behavior.

\subsection{Formal Representation: Variables, Predicates \& Premises}
\label{sec:chart_representation}

Charts are structured visual objects composed of axes, labels, legends, and data marks. VPP converts these components into a symbolic representation that supports logical reasoning and deterministic verification. Let $A$ denote the set of observable chart attributes
\[
A = \{\text{axes}, \text{ticks}, \text{labels}, \text{legends}, \text{data marks}\}.
\]
From these attributes we derive chart variables $\mathcal{V}$ corresponding to identifiable components such as axis labels, legend entries, tick values, and coordinate values.

Reasoning statements are expressed as predicates over these variables, e.g., \texttt{Value\_At(series, tick)}, \texttt{Legend\_Exist(series)}, and \texttt{Greater\_Than(value\_i, value\_j)}. Together, the variable set and predicate vocabulary define a closed reasoning space in which chart-grounded statements can be expressed and checked. This representation makes chart reasoning executable as compositions of verifiable logical statements grounded in observable visual components.

A key property of RealCQA-V2 is that each reasoning step is available in three aligned forms: \textbf{NLP}, a natural-language statement used for LVLM evaluation; \textbf{FOL}, a symbolic first-order predicate over chart-grounded variables; and \textbf{AST}, a graph-structured representation of the compositional reasoning chain. This multi-representation design makes the benchmark usable not only for language-based models, but also for symbolic, neuro-symbolic, and graph-based reasoning systems. Unlike free-form explanation datasets, RealCQA-V2 therefore provides a formally structured reasoning substrate in addition to natural-language supervision.

\subsection{Reasoning Chains and Provenance}
\label{sec:reasoning_chains}

Individual premises capture atomic reasoning steps, but answering a chart question typically requires multiple dependent steps. VPP therefore organizes premises into structured reasoning chains associated with each question. For a question $Q$, the reasoning chain is an ordered sequence
\[
P_Q = (p_1,p_2,\ldots,p_n)
\]
whose conjunction determines whether the hypothesis is satisfied.

Premise construction follows a deterministic chart interpretation pipeline,
\[
\text{Structural} \rightarrow \text{Data} \rightarrow \text{Reasoning} \rightarrow \text{Math},
\]
yielding four premise types. \textbf{Structural Premises (SP)} verify chart grammar such as axes, tick marks, and legends; \textbf{Data Premises (DP)} retrieve numerical values associated with chart coordinates; \textbf{Reasoning Premises (RP)} compare or relate numerical values; and \textbf{Math Premises (MP)} evaluate derived quantities such as ratios or aggregates. Later steps depend on earlier verified visual facts.

Chains can also be represented as directed acyclic graphs
\[
G_Q = (P_Q, E_Q),
\]
where nodes correspond to premises and edges encode logical dependencies. In addition to their natural-language form, premise chains are converted to first-order logic and parsed to generate abstract syntax trees (ASTs). We first generate candidate FOL translations from natural-language premise templates using GPT-4o, then apply human verification to ensure that the resulting predicate structure matches the intended chart-grounded semantics, with variables drawn from ChartInfo and predicates from the closed VPP vocabulary. Verified FOL sequences are subsequently parsed using \texttt{sympy} to produce ASTs that encode compositional dependency structure. These ASTs provide explicit provenance over the reasoning structure and make the benchmark compatible with graph-based verification and graph-isomorphism style reasoning tasks.

\begin{figure*}[t]
\centering
\includegraphics[width=\textwidth]{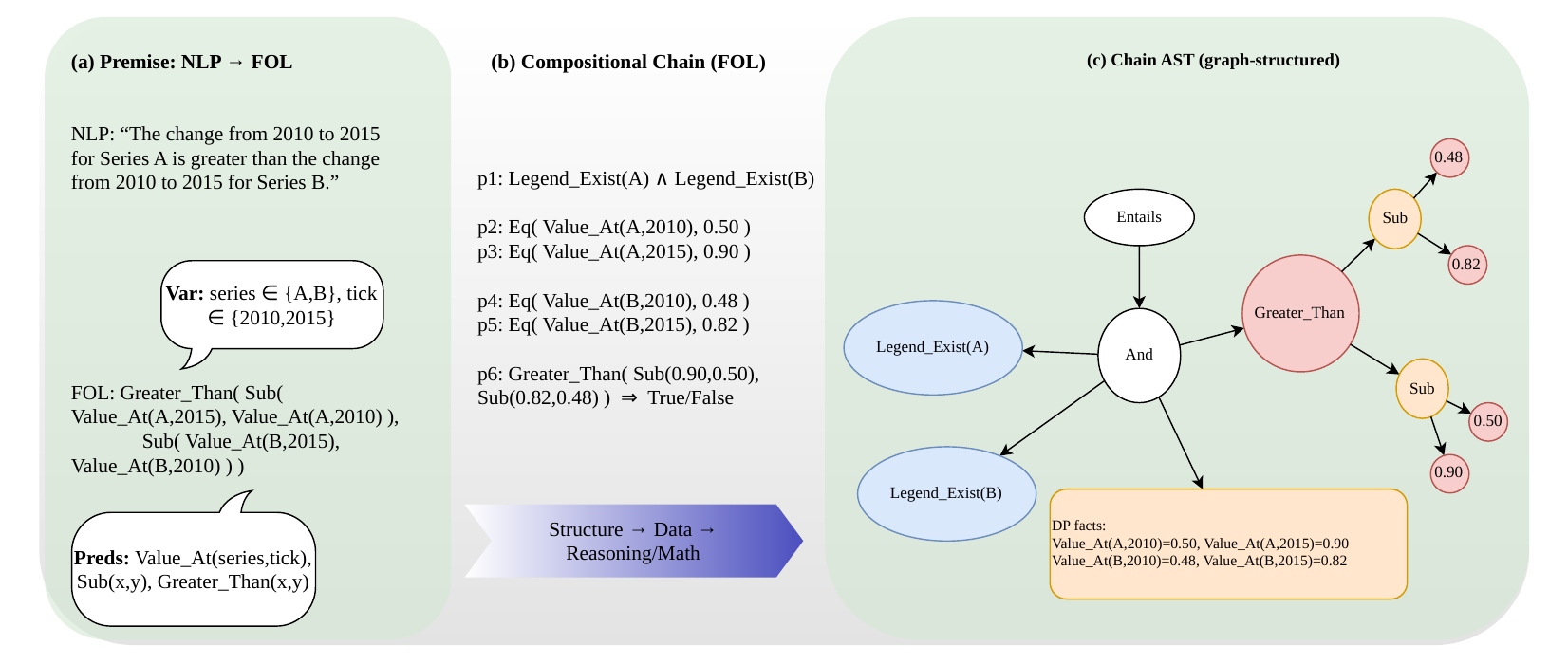}
\caption{
Multi-representation provenance in RealCQA-V2. Each reasoning chain is aligned across natural language, first-order logic, and abstract syntax tree (AST) forms, enabling evaluation with language-based, symbolic, and graph-structured reasoning systems.
}
\label{fig:chain}
\end{figure*}

Figure~\ref{fig:chain} illustrates this tri-level alignment between natural language, first-order logic, and AST structure. This representation is central to the novelty of RealCQA-V2: the benchmark is not limited to chart-grounded language statements, but also supports symbolic and graph-structured variants of visual reasoning.

\begin{figure*}[t]
\centering
\includegraphics[width=\textwidth]{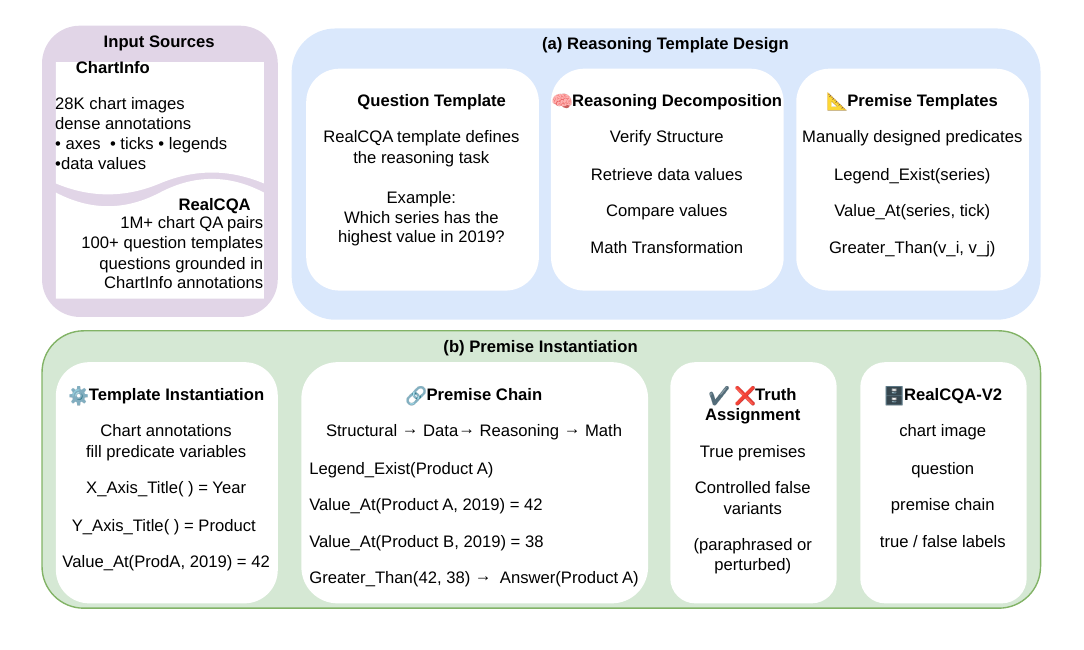}
\caption{
RealCQA-V2 construction. ChartInfo~\cite{davila2022icpr} provides chart images with component-level annotations, while RealCQA~\cite{ahmed2023realcqa} provides large-scale question templates grounded in those charts. For each question template, we manually design a reasoning decomposition and corresponding predicate-based premise templates; chart annotations then instantiate these templates into grounded premise chains with true/false labels.
}
\label{fig:vpp_dataset_pipeline}
\end{figure*}

\subsection{Dataset Construction Pipeline}
\label{sec:pipeline}

Figure~\ref{fig:vpp_dataset_pipeline} summarizes the construction of RealCQA-V2. The benchmark is derived through a staged transformation of ChartInfo annotations and RealCQA question templates into chart-grounded premise chains. ChartInfo~\cite{davila2022icpr} provides 28K scientific chart images with dense component-level annotations, including axes, tick values, legend entries, and data values. RealCQA~\cite{ahmed2023realcqa} builds on these charts to provide over 1M question--answer pairs generated from more than 100 question templates, with answers grounded in the underlying ChartInfo annotations.

For each RealCQA question template, we manually design a reasoning decomposition and the corresponding sequence of premise templates. These premise templates are the formal distillation of the original question pattern into predicate-based reasoning steps. Since multiple question strings may instantiate the same underlying reasoning template, each template defines a reusable compositional chain that can be instantiated across many chart instances.  The overall pipeline is:
\begin{center}
\small
\textit{ChartInfo annotations} $\rightarrow$
\textit{RealCQA templates} $\rightarrow$
\textit{Premise templates} $\rightarrow$\\
\textit{Variable instantiation} $\rightarrow$
\textit{Premise labeling}.
\end{center}
This pipeline preserves provenance from source annotations to premise labels, enabling reproducible construction of large-scale chart-grounded reasoning data.

\paragraph{Annotation Workforce and Protocol.}
Premise templates and chain decompositions were produced by a graduate-level annotation team operating under a structured review process. The team consisted of 7 graduate student annotators (M.S./Ph.D. programs in STEM fields such as computer science, data science, and information systems), supervised by a lead reviewer and a tiered review panel. Each question template was assigned to a primary annotator who authored the reasoning decomposition, the corresponding premise templates, and representative instantiations against ChartInfo variables. A lead reviewer performed template-level validation, while 3 additional reviewers audited coverage, consistency with chart semantics, and compatibility with the VPP premise taxonomy (SP/DP/RP/MP).

\paragraph{Quality Control and Consistency Checks.}
We apply quality control at the template, instantiation, and chain levels. At the template level, reviewers check that each premise is atomic and type-consistent, that premise variables refer to valid chart components, that the FOL form is well-typed under the closed variable/predicate sets, and that chain ordering respects dependency constraints. At the instantiation level, automated sanity checks ensure that all variable bindings exist in the ChartInfo annotation record and that numeric comparisons use values within valid chart ranges. At the chain level, we audit $\sim$5--10\% of chains per template for logical coherence and label correctness, with disagreements adjudicated by the lead reviewer. We will publish the annotation rubric, audit checklist, and validation scripts as part of the release.

\subsection{Premise Generation and Negative Construction}
\label{sec:premise_construction}

VPP represents reasoning steps as \emph{premises}, i.e., atomic logical statements derived from chart annotations and question templates. Premise templates are manually curated by experts for each RealCQA reasoning pattern and then instantiated with chart variables extracted from ChartInfo annotations. For each reasoning pattern, we define three base templates that generate grounded premise variants. These templates exist in both natural-language and symbolic first-order forms. To increase linguistic diversity while preserving semantics, instantiated premises are paraphrased using a T5-based paraphrasing model. The full template library, paraphrase variants, and symbolic forms are provided in the supplementary material.

Each instantiated premise yields one true statement and three controlled false variants. False premises are generated through systematic perturbations of axis or tick values, legend labels, numeric quantities, and comparison operators. This yields a consistent \textbf{3:1 False--True ratio}. Because the perturbations preserve syntactic form and chart-level plausibility, false premises remain semantically realistic and difficult to reject using superficial lexical cues alone. Since premises are instantiated directly from human chart annotations and question templates, every statement corresponds to a verifiable visual fact.

\paragraph{Artifact Mitigation for False Premises.}
False premises are generated via controlled perturbations---value swaps, operator flips, legend swaps, and range-consistent numeric perturbations---designed to preserve surface plausibility while breaking entailment. To reduce annotation artifacts and lexical shortcuts, we enforce matched template structure between true and false variants, apply paraphrase augmentation symmetrically across labels, and constrain perturbations so that format and variable bindings remain valid under the same predicate schema. As lightweight artifact checks, we evaluate premise-only baselines that ignore the chart image (e.g., a text-only classifier over premise strings and a label-prior baseline); these remain near chance on the evaluation split (51.3\%/48.7\% vs.\ a 25\%/75\% true/false prior), suggesting that strong performance requires chart-grounded verification rather than exploitation of superficial textual cues. Full artifact diagnostics and additional baselines are reported in the supplementary material.

\subsection{Verification Protocol}
\label{sec:verification}

VPP evaluates models by measuring whether they correctly verify premises and reasoning chains.

\paragraph{Premise Accuracy}
Premise accuracy measures local reasoning fidelity: whether a model can correctly judge an individual chart-grounded statement. Given chart $C$ and premise $p$, the model predicts whether the statement is entailed by the chart. Premise accuracy is the fraction of correctly verified premises.

\paragraph{VPP Accuracy (\(Acc_{VPP}\)).}
A reasoning chain is considered correct only if every premise in the chain is verified correctly. Chain accuracy therefore measures global reasoning consistency across the full sequence.
\[
ACC_{VPP} = \prod_i t_i.
\]
  
\paragraph{Depth of Correct Premises (DCP).}
To provide diagnostic insight into reasoning failures, we measure the depth of correct premises within failed chains. DCP quantifies how far a model progresses through a reasoning chain before failure, capturing partial reasoning competence within otherwise incorrect chains.
\[
DCP = \frac{1}{S - S_{correct}} 
\sum_{\substack{s=1 \\ \prod_{p=1}^{P_s} ACC_{s,p} = 0}}^{S} 
\left( \frac{1}{P_s} \sum_{p=1}^{P_s} AC C_{s,p} \right),
\]
where $S$ is the total number of chains and $S_{correct}$ those that are fully correct.

Together, these metrics evaluate both local premise correctness and global chain consistency, enabling the benchmark to reveal a measurable local--global reasoning gap.

\subsection{Dataset Snapshot and Split}
\label{sec:dataset_snapshot}

Table~\ref{tab:vpp_snapshot} summarizes the scale and composition of RealCQA-V2. The benchmark contains over \textbf{5.2M} training premises and \textbf{51K} evaluation premises derived from thousands of chart images. Premises follow a controlled \textbf{3:1 false-to-true ratio} and span four reasoning categories: structural verification, data retrieval, relational reasoning, and mathematical reasoning. To prevent image-level leakage, chart images are assigned exclusively to train or test splits.

\begin{table}[t]
\centering
\scriptsize
\caption{VPP dataset snapshot summarizing scale, reasoning composition, and train/test split statistics.}
\label{tab:vpp_snapshot}
\begin{tabular}{lcc}
\toprule
\textbf{Statistic} & \textbf{Train} & \textbf{Test} \\
\midrule
Premises & 5.21M & 51K \\
True / False & 1.33M / 3.88M & 13K / 38K \\
Charts (PMC IDs) & 15,098 & 4,039 \\
QA Reasoning Chains & 408K & 4,063 \\
Complete Chains & 408K & 2,197 \\
Split Chains & -- & 1,865 \\
False:True Ratio & \multicolumn{2}{c}{3:1} \\
\midrule
\multicolumn{3}{c}{\textbf{Premise Categories (Evaluation Split)}} \\
\midrule
DP (Data) & \multicolumn{2}{c}{19,757} \\
RP (Reasoning) & \multicolumn{2}{c}{15,507} \\
SP (Structural) & \multicolumn{2}{c}{14,351} \\
MP (Math) & \multicolumn{2}{c}{1,392} \\
\bottomrule
\end{tabular}
\end{table}

\section{Experimental Evaluation}
\label{sec:experiments}

We evaluate models on the VPP benchmark to probe complementary capabilities: \emph{local premise verification} and \emph{global chain consistency}. The goal of these experiments is to showcase RealCQA-V2 as a diagnostic benchmark for structured visual entailment across diverse reasoning settings and model families.

\subsection{Evaluation Setup}
\label{sec:exp_setup}

\textbf{Prompting modes.}
We evaluate models under two prompting modes.
\textit{Singleton} presents each $(\text{image}, \text{premise})$ pair independently and measures premise-level verification.
\textit{CoT} presents the full reasoning chain and requires the model to return one True/False judgment per premise.
This allows us to separate isolated premise verification from reasoning under explicit compositional context.

\textbf{Metrics.}
We report three metrics introduced in Section~\ref{sec:verification}:
(i) premise accuracy, which measures local reasoning fidelity;
(ii) $Acc_{VPP}$, which measures whether an entire reasoning chain is verified correctly; and
(iii) DCP, which measures the depth of correct premises within failed chains.
Prompt templates are provided in the supplementary material.

\textbf{Models.}
We evaluate both closed-source and open-source LVLM baselines:
\textbf{ChatGPT-4o}\footnote{https://openai.com/index/chatgpt},
\textbf{Gemini 2.5 Pro}\footnote{https://deepmind.google/technologies/gemini/},
and the \textbf{InternVL} family~\cite{Chen_2024_CVPR}, including InternVL2, InternVL2.5, and InternVL3 variants.
In addition, we include \textbf{IVL2.5-8b-FT-S}, a LoRA fine-tuned variant trained on singleton premise supervision, as a simple aligned baseline.
This setup is intended to showcase the benchmark across model families rather than to optimize a new method.

\subsection{Premise-Level Verification}
\label{sec:premise_evaluation}

Table~\ref{tab:premise_accuracy_by_model} reports premise-level accuracy across models.
These results show that atomic premise verification is already a non-trivial task, with substantial variation across model families, label polarity, and prompting mode.

Gemini~2.5 achieves the strongest overall premise accuracy on the full evaluation set (\textbf{76.33}\%), with relatively balanced performance across true and false premises.
ChatGPT-4o is competitive but lower overall, while the InternVL family exhibits sharper asymmetries.
For example, InternVL2-8b attains strong false-premise rejection (\textbf{92.52}\%) but performs poorly on true premises (27.71\%), whereas InternVL3-14b shows the opposite pattern, reaching the highest true-premise accuracy (\textbf{83.60}\%) but substantially weaker false-premise discrimination.
These complementary error modes indicate that premise verification depends not only on visual recognition, but also on calibration across premise types and truth values.

A second pattern is the effect of compositional context.
For most open-source models, CoT prompting improves premise accuracy on the chain subset, suggesting that neighboring premises help constrain local judgments.
However, this gain is not universal: Gemini~2.5 performs best in singleton mode on the full evaluation set, indicating that explicit chains do not always improve local verification for already strong closed-source systems.
This contrast is precisely the kind of behavior VPP is designed to expose.

The singleton fine-tuned baseline illustrates this point.
IVL2.5-8b-FT-S improves premise accuracy on the CoT subset (\textbf{81.23}\%) relative to zero-shot , but remains far from the strongest models on the full 51K evaluation set.
This shows that aligned local supervision can improve atomic verification, but does not by itself solve the broader challenge of compositional reasoning.

Overall, premise-level results establish that VPP is a meaningful local verification benchmark.
At the same time, they also foreshadow the central finding of this paper: strong local premise accuracy does not necessarily translate into coherent full-chain reasoning.

    \begin{table}[t]
\centering
\scriptsize
\caption{Premise-level accuracy (\%) and total premises evaluated. Best result per column in \textbf{bold}.}
\label{tab:premise_accuracy_by_model}
\begin{tabular}{lcccc}
\toprule
\textbf{Model} 
& \textbf{CoT (12k)} 
& \textbf{All (51k)} 
& \textbf{True (13k)} 
& \textbf{False (38k)} \\
\midrule
CGPT-4o         & 74.28 & 66.16 & 69.85 & 64.89 \\
Gemini 2.5      & 61.08 & \textbf{76.33} & 81.11 & 74.69 \\
IVL2-8b         & 67.29 & 75.93 & 27.71 & \textbf{92.52} \\
IVL2.5-8b       & 78.56 & 60.92 & 69.36 & 58.02 \\
IVL2.5-FT-S     & \textbf{81.23} & 61.67 & 65.43 & 60.38 \\
IVL3-8b         & 71.12 & 69.19 & 60.24 & 72.27 \\
IVL3-9b         & 84.69 & 72.93 & 47.64 & 81.64 \\
IVL3-14b        & 66.45 & 58.13 & \textbf{83.60} & 49.36 \\
\bottomrule
\end{tabular}
\end{table}

\subsection{Chain-Level Verification}
\label{sec:chain_evaluation}

We now evaluate whether models can preserve logical consistency across complete reasoning chains.
Table~\ref{tab:chain_reasoning_model_results} reports premise accuracy, full-chain accuracy ($Acc_{VPP}$), and DCP under both Singleton and CoT prompting.

The clearest finding is that \emph{local correctness and global consistency are not the same capability}.
Closed-source models such as ChatGPT-4o and Gemini~2.5 achieve high premise accuracy under CoT prompting (81.80\% and 78.34\%, respectively) and correspondingly strong DCP (79.60 and 75.72), yet their full-chain accuracy remains much lower (11.25\% and 24.03\%).
In other words, these models often verify many individual premises correctly, but fail to maintain coherence across the entire chain.
This local--global discrepancy is the central failure mode revealed by VPP.

Among the open-source baselines, performance varies sharply.
InternVL3-9b is the strongest zero-shot model, reaching \textbf{79.15}\% $Acc_{VPP}$ and \textbf{87.88}\% DCP in CoT mode.
Several InternVL variants also benefit substantially from explicit chain prompting: for example, InternVL2.5-8b improves from 14.56\% to 47.28\% $Acc_{VPP}$ when moving from Singleton to CoT evaluation.
These gains indicate that some models can exploit structured external chains effectively, even when their singleton premise verification is weaker.

The singleton fine-tuned baseline provides an informative contrast.
IVL2.5-8b-FT-S, despite being trained only on singleton premise supervision, improves from 4.82\% to 42.66\% $Acc_{VPP}$ when evaluated with CoT prompting.
This suggests that local supervision can help activate latent chaining behavior, but still falls short of fully aligned compositional reasoning.
Importantly, this result supports the benchmark claim rather than a method claim: VPP distinguishes between models that can classify individual statements and those that can sustain logical consistency over an entire reasoning sequence.

Taken together, the chain-level results demonstrate the main value of RealCQA-V2.
The benchmark does not merely measure whether a model can read charts or classify isolated statements; it reveals whether local entailments compose into globally valid reasoning.
This makes $Acc_{VPP}$ and DCP necessary complements to premise-level accuracy.

\begin{table}[t]
\centering
\scriptsize
\caption{Chain-level performance across models and prompt modes. 
Arrows indicate CoT gains over Singleton.}
\label{tab:chain_reasoning_model_results}

\begin{tabular}{l l c c c c c}
\toprule
\textbf{Model} & \textbf{Mode}
& \textbf{PremAcc}
& \textbf{Acc$_{\text{VPP}}$}
& \textbf{DCP}
& $\Delta$Acc$_{\text{VPP}}$
& $\Delta$DCP \\
\midrule

CGPT-4o 
& CoT        & 81.80 & 11.25 & 79.60 & \downarrowred 2.63 & \uparrowgreen 3.24 \\
& Singleton  & 67.81 & 13.88 & 76.36 &  &  \\
\midrule

Gemini 2.5 
& CoT        & 78.34 & 24.03 & 75.72 & \downarrowred 8.42 & \uparrowgreen 5.70 \\
& Singleton  & 74.36 & 32.45 & 81.42 &  &  \\
\midrule

InternVL2-8b
& CoT        & 82.52 & 17.61 & 82.14 & \uparrowgreen 17.61 & \uparrowgreen 33.28 \\
& Singleton  & 41.28 & 0.00  & 48.86 &  &  \\
\midrule

InternVL2.5-8b
& CoT        & 88.07 & 47.28 & 85.74 & \uparrowgreen 32.72 & \uparrowgreen 7.17 \\
& Singleton  & 69.26 & 14.56 & 78.57 &  &  \\
\midrule

InternVL2.5-8b-FT-S
& CoT        & 85.63 & 42.66 & 83.31 & \uparrowgreen 37.84 & \uparrowgreen 10.78 \\
& Singleton  & 65.01 & 4.82  & 72.53 &  &  \\
\midrule


InternVL3-8b
& CoT        & 84.11 & 39.01 & 81.43 & \uparrowgreen 30.00 & \uparrowgreen 5.57 \\
& Singleton  & 66.31 & 9.01  & 75.86 &  &  \\
\midrule

InternVL3-9b
& CoT        & 92.15 & 79.15 & 87.88 & \uparrowgreen 76.56 & \uparrowgreen 23.97 \\
& Singleton  & 54.41 & 2.59  & 63.91 &  &  \\
\midrule

InternVL3-14b
& CoT        & 80.42 & 11.68 & 81.48 & \downarrowred 22.95 & \downarrowred 0.86 \\
& Singleton  & 75.42 & 34.63 & 82.34 &  &  \\

\bottomrule
\end{tabular}
\end{table}

\subsection{Transfer to Final-Answer Chart QA}
\label{sec:vqa_comparison}

To assess whether structured premise supervision also benefits conventional answer prediction, we evaluate transfer on the original RealCQA task (Table~\ref{tab:vqa_summary}). Structured supervision improves final-answer accuracy for smaller (pre-LLM) chart reasoning models, suggesting that explicit premise-level training can provide useful inductive bias for answer prediction for smaller parameter models. 
However, this transfer setting is secondary to our main focus.
The central contribution of RealCQA-V2 is not a new answering model, but a benchmark that exposes and measures structured visual reasoning directly.

Additional analyses of premise-type performance, per-image distributions, and AST-level reasoning behavior are provided in the supplementary material. These analyses further illustrate the diagnostic range of the benchmark, but are omitted from the main paper for clarity.

    \begin{table}[t]
\centering
\scriptsize
\caption{Condensed VQA accuracy (\%) across key evaluation axes. 
Full breakdown in Supplementary}
\label{tab:vqa_summary}

\begin{tabular}{l c c c c c}
\toprule
\textbf{Model} 
& \textbf{Total} 
& \textbf{Binary} 
& \textbf{Structural} 
& \textbf{Reasoning} 
& \textbf{Chart Avg} \\
\midrule

VL-T5                    & 31.06 & 52.75 & 43.52 & 29.37 & 30.2 \\
CRCT                     & 18.80 & 18.07 & 14.98 & 19.60 & 18.8 \\
UniChart                 & 26.75 & 51.53 & 21.40 & 27.64 & 26.8 \\
ViT+T5 (Matcha)          & 25.97 & 52.54 & 19.85 & 27.71 & 26.6 \\
ViT+T5 (RealCQA)         & 32.10 & 56.19 & 42.41 & 30.89 & 36.6 \\

\textbf{ViT+T5 (RealCQA-V2)} 
& \textbf{44.62} 
& \textbf{67.95} 
& \textbf{83.89} 
& \textbf{38.84} 
& \textbf{54.8} \\
\bottomrule
\end{tabular}
\end{table}

\section*{Conclusion}

We introduced \textbf{RealCQA-V2}, a large-scale benchmark for \emph{structured visual entailment} over real scientific charts, and formalized its evaluation through \textbf{Visual Premise Proving (VPP)}. By decomposing chart question answering into atomic, chart-grounded premises with aligned \textbf{natural-language}, \textbf{first-order logic}, and \textbf{AST} representations, the benchmark enables reproducible verification of intermediate reasoning steps beyond answer-only accuracy. Our experiments reveal a clear \emph{local--global reasoning gap}: models can verify many individual premises correctly yet still fail to maintain coherence across full reasoning chains. We release RealCQA-V2 to support future work on verifiable multimodal reasoning.

\clearpage
\appendix
\setcounter{page}{1}

\section*{Appendix Overview}
\addcontentsline{toc}{section}{Appendix Overview}

This appendix provides supplementary material supporting the results
presented in the main paper. It includes qualitative reasoning examples,
additional distributional analyses, and extended benchmark evaluations.

The appendix is organized as follows:

\begin{itemize}

\item \textbf{Appendix A: Qualitative Failure Analysis}  
Illustrative reasoning traces from the Visual Premise Proving (VPP)
evaluation pipeline. These examples visualize how models process
structural premises (SP), data premises (DP), and relational premises (RP),
and highlight common failure modes in multimodal reasoning.

\item \textbf{Appendix B: Distributional Performance Analysis}  
Additional plots analyzing prediction accuracy distributions across
different structural levels of the VPP framework, including
image-level aggregation, premise tags, reasoning chains (ASTs),
and premise source types.

\item \textbf{Appendix C: Full VQA Benchmark Results}  
Comprehensive benchmark results for the VQA evaluation, including
performance across answer categories, question types, and chart
modalities.

\item \textbf{Appendix D: Dataset Details}  
Dense Chart parsing, Chart representation as FOL, predicates, premise creation, premise templates, dataset statistics, release form, annotator details. 

\end{itemize}

\clearpage

\section{Qualitative Failure Analysis}
\label{app:qualitative}

This section presents qualitative reasoning traces from the
\textbf{Visual Premise Proving (VPP)} evaluation.
Each example visualizes the reasoning pipeline for a single question
instance grounded in the source chart.

For each case we display:

\begin{itemize}
\item the original chart image,
\item the sequence of \textbf{structural premises (SP)},
      \textbf{data premises (DP)}, and the final
      \textbf{relational premise (RP)},
\item model predictions from
      \textbf{InternVL3-9B}, \textbf{Gemini 2.5 Pro}, and \textbf{GPT-4o}.
\end{itemize}

Each premise is evaluated independently as a True/False statement.
Correct model predictions are marked by \cmark\ and incorrect predictions
by \xmark.

These examples illustrate several common reasoning failure modes,
including perceptual grounding errors, incorrect data extraction,
and multi-hop relational reasoning mistakes.

\clearpage

\figsep{1}

\begin{center}
\includegraphics[width=0.9\linewidth]{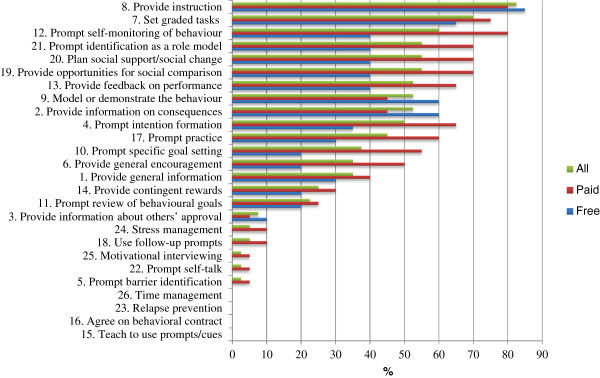}
\end{center}

\small
\setlength{\tabcolsep}{2pt}
\renewcommand{\arraystretch}{1.05}

\begin{longtable}{p{2.5cm}p{3.5cm}ccc}
\toprule
Tag & Premise & InternVL3-9B & Gemini2.5Pro & GPT4o \\
\midrule
\endhead
SP0 & The type of chart is horizontal bar. & \cmark & \cmark & \cmark \\
SP1 & The dependent axis is labeled as \%. & \cmark & \cmark & \xmark \\
SP3 & The dependent axis ranges from a minimum of 0 to a maximum of 90 in \%. & \cmark & \cmark & \cmark \\
SP7 & Tick marks corresponding to specified \% values are present on the dependent axis. & \cmark & \cmark & \cmark \\
SP8 & The chart contains a legend that differentiates between the 3 data series. & \xmark & \cmark & \cmark \\
SP9 & Each data series in the legend corresponds to a unique representation on the chart (e.g., color, pattern, line type) and has the labels ['Free', 'Paid', 'All']. & \cmark & \cmark & \cmark \\
\midrule
DP\_leg\_i & The axis of \% for the data series Free has values at points (13. Provide feedback on performance) & \xmark & \cmark & \xmark \\
DP\_leg\_j & The axis of \% for the data series Paid has values at points (26. Time management) & \xmark & \xmark & \xmark \\
\midrule
RP\_68 & The difference in \% for Free between x-ticks 13. Provide feedback on performance and 26. Time management is less than the difference for Paid between x-ticks 13. Provide feedback on performance and 26. Time management & \xmark & \cmark & \cmark \\
\bottomrule
\end{longtable}

\figsep{2}

\begin{center}
\includegraphics[width=0.9\linewidth]{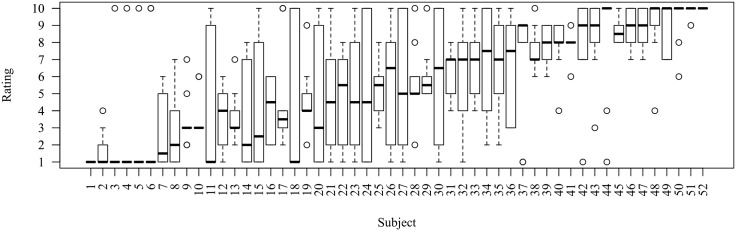}
\end{center}

\small
\setlength{\tabcolsep}{2pt}
\renewcommand{\arraystretch}{1.05}

\begin{longtable}{p{2.5cm}p{3.5cm}ccc}
\toprule
Tag & Premise & InternVL3-9B & Gemini2.5Pro & GPT4o \\
\midrule
\endhead
SP0 & The type of chart is vertical box.  & \xmark & \cmark & \cmark \\
SP1 & The dependent axis is labeled as Rating.  & \cmark & \cmark & \cmark \\
SP2 & The independent axis is labeled as Subject.  & \cmark & \cmark & \cmark \\
SP3 & The dependent axis ranges from a minimum of 1 to a maximum of 10 in Rating.  & \cmark & \cmark & \cmark \\
SP4 & The independent axis ranges from a minimum of 1 to a maximum of 52 in Subject.  & \xmark & \cmark & \cmark \\
SP6 & Tick marks corresponding to specified Subject values are present on the independent axis.  & \xmark & \cmark & \cmark \\
SP7 & Tick marks corresponding to specified Rating values are present on the dependent axis.  & \cmark & \cmark & \cmark \\
\midrule
DP\_TQ\_exist\_i & Upper quartile data exists for Rating at 10  & \xmark & \cmark & \cmark \\
DP\_TQ\_exist\_j & At x-tick 15, the upper quartile of Rating is defined  & \xmark & \cmark & \cmark \\
DP\_TQ\_val\_i & The 75th percentile value for Rating at 10 is 3.11  & \xmark & \xmark & \xmark \\
DP\_TQ\_val\_j & At x-tick 15, the third quartile of Rating equals 8.02  & \xmark & \xmark & \xmark \\
\midrule
RP\_167 & The upper quartile of Rating at x-tick 10 is less than that at x-tick 15  & \xmark & \xmark & \cmark \\
\bottomrule
\end{longtable}

\figsep{3}

\begin{center}
\includegraphics[width=0.9\linewidth]{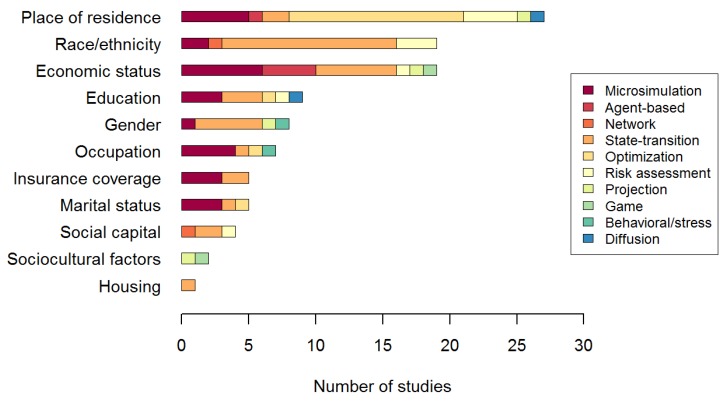}
\end{center}

\small
\setlength{\tabcolsep}{2pt}
\renewcommand{\arraystretch}{1.05}

\begin{longtable}{p{2.5cm}p{3.5cm}ccc}
\toprule
Tag & Premise & InternVL3-9B & Gemini2.5Pro & GPT4o \\
\midrule
\endhead
SP0 & The type of chart is horizontal bar.& \cmark & \cmark & \cmark \\
SP1 & The dependent axis is labeled as Number of studies.& \cmark & \xmark & \xmark \\
SP3 & The dependent axis ranges from a minimum of 0 to a maximum of 30 in Number of studies.& \cmark & \cmark & \xmark \\
SP7 & Tick marks corresponding to specified Number of studies values are present on the dependent axis.& \cmark & \cmark & \cmark \\
SP8 & The chart contains a legend that differentiates between the 10 data series.& \cmark & \cmark & \cmark \\
SP9 & Each data series in the legend corresponds to a unique representation on the chart (e.g., color, pattern, line type) and has the labels ['Microsimulation', 'Agent based', 'Network', 'State transition', 'Optimization', 'Risk assesment', 'Behavioral/stress', 'Projection', 'Diffusion', 'Game'].& \cmark & \xmark & \xmark \\
DP\_leg\_ii & Value in the chart plot area exists at (Race/ ethinicity) for the axis called Number of studies for the data series Risk assesment& \xmark & \xmark & \cmark \\
DP\_leg\_ij & Value in the chart plot area exists at (Insurance coverage) for the axis called Number of studies for the data series Diffusion& \xmark & \xmark & \xmark \\
RP\_68 & The difference in Number of studies for Risk assesment between x-ticks Race/ ethinicity and Insurance coverage is greater than the difference for Diffusion between x-ticks Race/ ethinicity and Insurance coverage& \cmark & \xmark & \cmark \\
\bottomrule
\end{longtable}

\figsep{4}

\begin{center}
\includegraphics[width=0.9\linewidth]{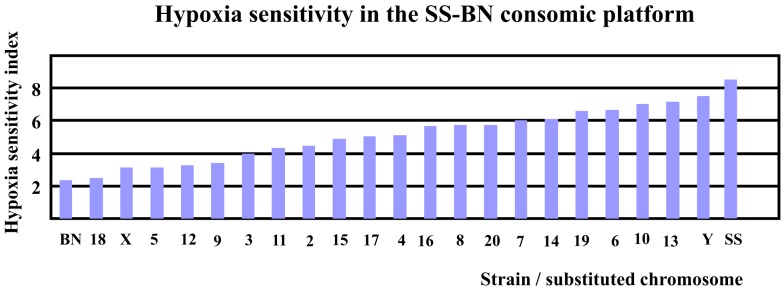}
\end{center}

\small
\setlength{\tabcolsep}{2pt}
\renewcommand{\arraystretch}{1.05}

\begin{longtable}{p{2.5cm}p{3.5cm}ccc}
\toprule
Tag & Premise & InternVL3-9B & Gemini2.5Pro & GPT4o \\
\midrule
\endhead
SP0 & The type of chart is vertical bar.& \cmark & \cmark & \cmark \\
SP1 & The dependent axis is labeled as Hypoxia sensitivity index.& \cmark & \cmark & \cmark \\
SP2 & The independent axis is labeled as Strain / substituted chromosome.& \cmark & \cmark & \cmark \\
SP3 & The dependent axis ranges from a minimum of 2 to a maximum of 8 in Hypoxia sensitivity index.& \cmark & \xmark & \xmark \\
SP5 & The independent axis is categorical with the labels ['BN', 'SS', 'Y', 13, 10, 6, 19, 14, 20, 7, 8, 4, 17, 16, 15, 2, 11, 3, 9, 12, 5, 'X', 18].& \cmark & \cmark & \cmark \\
SP6 & Tick marks corresponding to specified Strain / substituted chromosome values are present on the independent axis.& \cmark & \xmark & \cmark \\
SP7 & Tick marks corresponding to specified Hypoxia sensitivity index values are present on the dependent axis.& \cmark & \cmark & \cmark \\
DP\_i & Value in the chart plot area exists at (X) for the axis called Hypoxia sensitivity index& \xmark & \cmark & \cmark \\
DP\_j & For the axis of Hypoxia sensitivity index, there are valid plot values corresponding to (5)& \cmark & \cmark & \cmark \\
RP\_59 & For Hypoxia sensitivity index, the value at X is lower than the value at 5& \xmark & \cmark & \xmark \\
\bottomrule
\end{longtable}

\figsep{5}

\begin{center}
\includegraphics[width=0.9\linewidth]{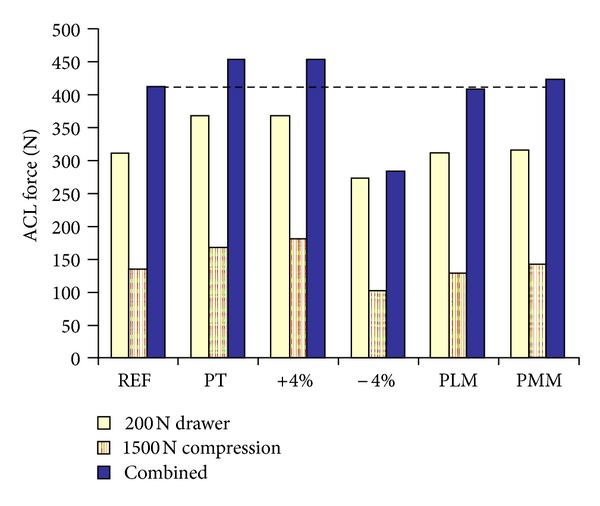}
\end{center}

\small
\setlength{\tabcolsep}{2pt}
\renewcommand{\arraystretch}{1.05}

\begin{longtable}{p{2.5cm}p{3.5cm}ccc}
\toprule
Tag & Premise & InternVL3-9B & Gemini2.5Pro & GPT4o \\
\midrule
\endhead
SP0 & The type of chart is vertical bar.& \cmark & \cmark & \cmark \\
SP1 & The dependent axis is labeled as ACL force (N).& \cmark & \cmark & \cmark \\
SP3 & The dependent axis ranges from a minimum of 0 to a maximum of 500 in ACL force (N).& \cmark & \cmark & \cmark \\
SP7 & Tick marks corresponding to specified ACL force (N) values are present on the dependent axis.& \cmark & \cmark & \cmark \\
SP8 & The chart contains a legend that differentiates between the 3 data series.& \cmark & \cmark & \cmark \\
SP9 & Each data series in the legend corresponds to a unique representation on the chart (e.g., color, pattern, line type) and has the labels ['200 N drawer', 'Combined', '1500 N compression'].& \xmark & \xmark & \xmark \\
DP\_leg\_ii & For the axis of ACL force (N), there are valid plot values corresponding to (REF) of the data series 1500 N compression& \cmark & \cmark & \cmark \\
DP\_leg\_ij & Value in the chart plot area exists at (-4\%) for the axis called ACL force (N) for the data series 1500 N compression& \xmark & \cmark & \cmark \\
RP\_62 & The value of ACL force (N) for 1500 N compression at x-tick REF is greater than that at x-tick -4\%& \cmark & \cmark & \xmark \\
\bottomrule
\end{longtable}

\figsep{6}

\begin{center}
\includegraphics[width=0.9\linewidth]{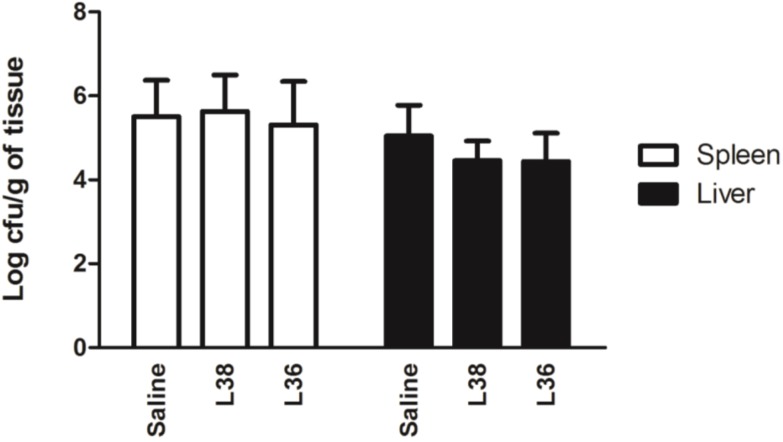}
\end{center}

\small
\setlength{\tabcolsep}{2pt}
\renewcommand{\arraystretch}{1.05}

\begin{longtable}{p{2.5cm}p{3.5cm}ccc}
\toprule
Tag & Premise & InternVL3-9B & Gemini2.5Pro & GPT4o \\
\midrule
\endhead
SP0 & The type of chart is vertical bar.& \cmark & \cmark & \cmark \\
SP1 & The dependent axis is labeled as Log cfu/g of tissue.& \cmark & \cmark & \cmark \\
SP3 & The dependent axis ranges from a minimum of 0 to a maximum of 8 in Log cfu/g of tissue.& \cmark & \cmark & \cmark \\
SP7 & Tick marks corresponding to specified Log cfu/g of tissue values are present on the dependent axis.& \xmark & \cmark & \cmark \\
SP8 & The chart contains a legend that differentiates between the 2 data series.& \cmark & \cmark & \cmark \\
SP9 & Each data series in the legend corresponds to a unique representation on the chart (e.g., color, pattern, line type) and has the labels ['Spleen', 'Liver'].& \cmark & \cmark & \cmark \\
DP\_leg\_ii & Value in the chart plot area exists at (L36) for the axis called Log cfu/g of tissue for the data series Liver& \xmark & \cmark & \cmark \\
DP\_leg\_ij & Value in the chart plot area exists at (Saline) for the axis called Log cfu/g of tissue for the data series Liver& \cmark & \cmark & \cmark \\
RP\_62 & The value of Log cfu/g of tissue for Liver at x-tick L36 is less than that at x-tick Saline& \cmark & \cmark & \xmark \\
\bottomrule
\end{longtable}

\figsep{7}

\begin{center}
\includegraphics[width=0.9\linewidth]{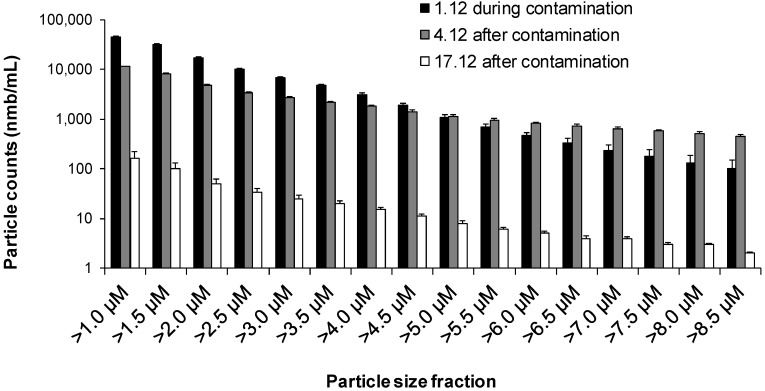}
\end{center}

\small
\setlength{\tabcolsep}{2pt}
\renewcommand{\arraystretch}{1.05}

\begin{longtable}{p{2.5cm}p{3.5cm}ccc}
\toprule
Tag & Premise & InternVL3-9B & Gemini2.5Pro & GPT4o \\
\midrule
\endhead
SP0 & The type of chart is vertical bar.& \cmark & \cmark & \cmark \\
SP1 & The dependent axis is labeled as Particle counts (nmb/mL).& \cmark & \cmark & \cmark \\
SP2 & The independent axis is labeled as Particle size fraction.& \cmark & \cmark & \cmark \\
SP3 & The dependent axis ranges from a minimum of 1 to a maximum of 100000.0 in Particle counts (nmb/mL).& \cmark & \cmark & \cmark \\
SP5 & The independent axis is categorical with the labels ['>1.0 uM', '>1.5 uM', '>2.0 uM', '>2.5 uM', '>3.0 uM', '>3.5 uM', '>4.0 uM', '>4.5 uM', '>5.0 uM', '>5.5 uM', '>6.0 uM', '>6.5 uM', '>7.0 uM', '>7.5 uM', '>8.0 uM', '>8.5 uM'].& \cmark & \cmark & \cmark \\
SP6 & Tick marks corresponding to specified Particle size fraction values are present on the independent axis.& \cmark & \xmark & \cmark \\
SP7 & Tick marks corresponding to specified Particle counts (nmb/mL) values are present on the dependent axis.& \xmark & \cmark & \cmark \\
SP8 & The chart contains a legend that differentiates between the 3 data series.& \cmark & \cmark & \cmark \\
SP9 & Each data series in the legend corresponds to a unique representation on the chart (e.g., color, pattern, line type) and has the labels ['1.12 during contamination', '4.12 after contamination', '17.12 after contamination'].& \xmark & \cmark & \cmark \\
DP\_leg\_ii & The axis of Particle counts (nmb/mL) for the data series 17.12 after contamination has values at points (>4.5 uM)& \xmark & \cmark & \xmark \\
DP\_leg\_ij & The axis of Particle counts (nmb/mL) for the data series 17.12 after contamination has values at points (>6.5 uM)& \cmark & \cmark & \cmark \\
RP\_62 & The difference of values of Particle counts (nmb/mL) for 17.12 after contamination at x-tick >4.5 uM and x-tick >6.5 uM is greater than zero& \xmark & \cmark & \cmark \\
\bottomrule
\end{longtable}

\figsep{8}

\begin{center}
\includegraphics[width=0.9\linewidth]{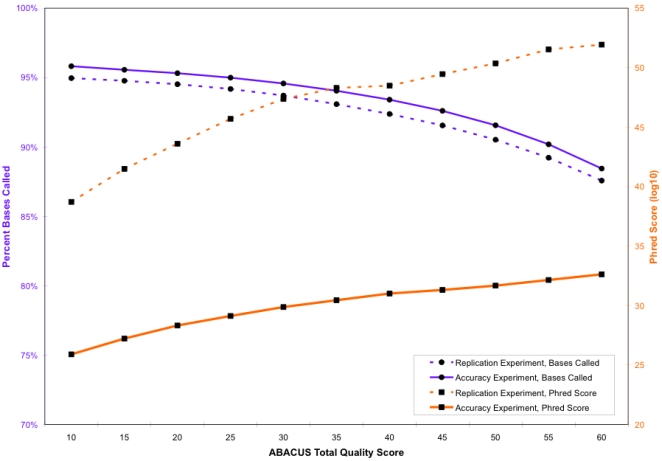}
\end{center}

\small
\setlength{\tabcolsep}{2pt}
\renewcommand{\arraystretch}{1.05}

\begin{longtable}{p{2.5cm}p{3.5cm}ccc}
\toprule
Tag & Premise & InternVL3-9B & Gemini2.5Pro & GPT4o \\
\midrule
\endhead
SP0 & The type of chart is line.& \cmark & \cmark & \cmark \\
SP1 & The dependent axis is labeled as Percent Bases Called.& \cmark & \cmark & \cmark \\
SP2 & The independent axis is labeled as Phred Score (log10).& \cmark & \xmark & \xmark \\
SP4 & The independent axis ranges from a minimum of 10 to a maximum of 60 in Phred Score (log10).& \cmark & \xmark & \xmark \\
SP6 & Tick marks corresponding to specified Phred Score (log10) values are present on the independent axis.& \cmark & \xmark & \xmark \\
SP7 & Tick marks corresponding to specified Percent Bases Called values are present on the dependent axis.& \cmark & \cmark & \cmark \\
SP8 & The chart contains a legend that differentiates between the 4 data series.& -- & \cmark & \cmark \\
SP9 & Each data series in the legend corresponds to a unique representation on the chart (e.g., color, pattern, line type) and has the labels ['Replication Experiment, Bases Called', 'Accuracy Experiment, Bases Called', 'Replication Experiment, Phred Score', 'Accuracy Experiment, Phred Score'].& \cmark & \cmark & \cmark \\
DP\_leg\_ii & For the axis of Percent Bases Called, there are valid plot values corresponding to (10) of the data series Accuracy Experiment, Bases Called& \xmark & \cmark & \cmark \\
DP\_leg\_ij & Value in the chart plot area exists at (20) for the axis called Percent Bases Called for the data series Accuracy Experiment, Phred Score& \xmark & \xmark & \xmark \\
RP\_68 & The gap in Percent Bases Called values for Accuracy Experiment, Bases Called (10 vs 20) does not exceed the gap for Accuracy Experiment, Phred Score (10 vs 20)& \xmark & \cmark & \cmark \\
\bottomrule
\end{longtable}

\figsep{9}
==========================================

\begin{center}
\includegraphics[width=0.9\linewidth]{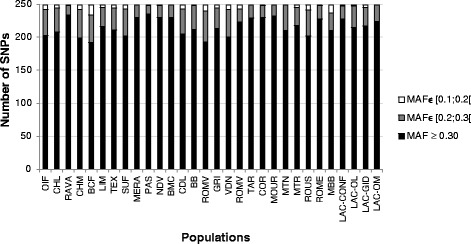}
\end{center}

\small
\setlength{\tabcolsep}{2pt}
\renewcommand{\arraystretch}{1.05}

\begin{longtable}{p{2.5cm}p{3.5cm}ccc}
\toprule
Tag & Premise & InternVL3-9B & Gemini2.5Pro & GPT4o \\
\midrule
\endhead
SP0 & The type of chart is vertical bar.& \cmark & \cmark & \cmark \\
SP1 & The dependent axis is labeled as Number of SNPs.& \cmark & \cmark & \cmark \\
SP2 & The independent axis is labeled as Populations.& \cmark & \cmark & \cmark \\
SP3 & The dependent axis ranges from a minimum of 0 to a maximum of 250 in Number of SNPs.& \cmark & \cmark & \cmark \\
SP5 & The independent axis is categorical with the labels ['LAC-CONF', 'LAC-OM', 'LAC-GID', 'LAC-OL', 'MBB', 'ROME', 'ROUS', 'MTR', 'MTN', 'MOUR', 'COR', 'TAR', 'ROMV', 'VDN', 'GRI', 'ROMV', 'BB', 'CDL', 'BMC', 'NDV', 'PAS', 'MERA', 'SUF', 'TEX', 'LIM', 'BCF', 'CHM', 'RAVA', 'CHIL', 'OIF'].& \xmark & \xmark & \xmark \\
SP6 & Tick marks corresponding to specified Populations values are present on the independent axis.& -- & \cmark & \cmark \\
SP7 & Tick marks corresponding to specified Number of SNPs values are present on the dependent axis.& \xmark & \cmark & \cmark \\
SP8 & The chart contains a legend that differentiates between the 3 data series.& \cmark & \cmark & \cmark \\
DP\_leg\_ii & For the axis of Number of SNPs, there are valid plot values corresponding to (ROMV) of the data series MAF \textbackslash geq 0.30& \xmark & \cmark & \cmark \\
DP\_leg\_ij & Value in the chart plot area exists at (COR) for the axis called Number of SNPs for the data series MAF \textbackslash geq 0.30& \xmark & \cmark & \cmark \\
RP\_62 & The difference of values of Number of SNPs for MAF \textbackslash geq 0.30 at x-tick COR and x-tick ROMV is greater than zero& \xmark & \cmark & \cmark \\
\bottomrule
\end{longtable}

\clearpage
\FloatBarrier

\section{Extended Quantitative Analysis}
To better understand model behavior beyond aggregate accuracy metrics,
we analyze how prediction accuracy varies across different structural
components of the Visual Premise Proving (VPP) evaluation pipeline.
Instead of reporting only global averages, we compute accuracy after
grouping predictions along specific structural axes of the dataset.

For each grouping axis (e.g., image, premise tag, reasoning chain),
all predictions belonging to that unit are aggregated to compute a
local accuracy score. The resulting set of group-level accuracies
forms a distribution that reflects how model performance varies
across different reasoning contexts.

\begin{figure*}[h]
\centering
\includegraphics[width=0.85\linewidth]{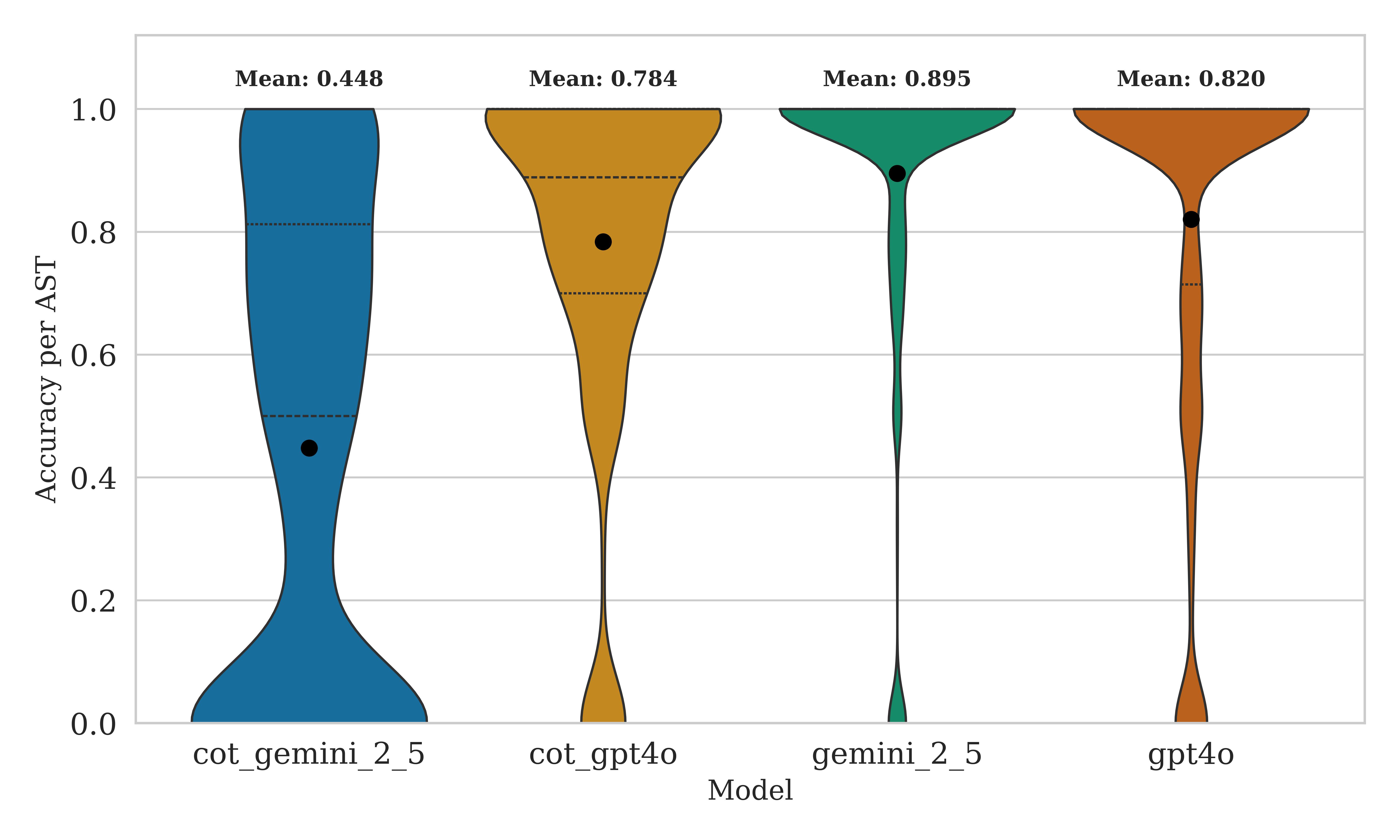}
\caption{
\textbf{Accuracy distribution across images.}
Each violin represents the distribution of per-image accuracy,
computed over all premises derived from the same chart.
While median accuracy remains relatively high across models,
the long lower tails indicate that a subset of charts produces
substantially more reasoning errors.
This suggests that difficulty is highly dependent on specific
chart structures rather than uniformly distributed across the dataset.
}
\label{fig:accuracy_image_distribution}
\end{figure*}

\begin{figure*}[ht]
\centering
\includegraphics[width=0.85\linewidth]{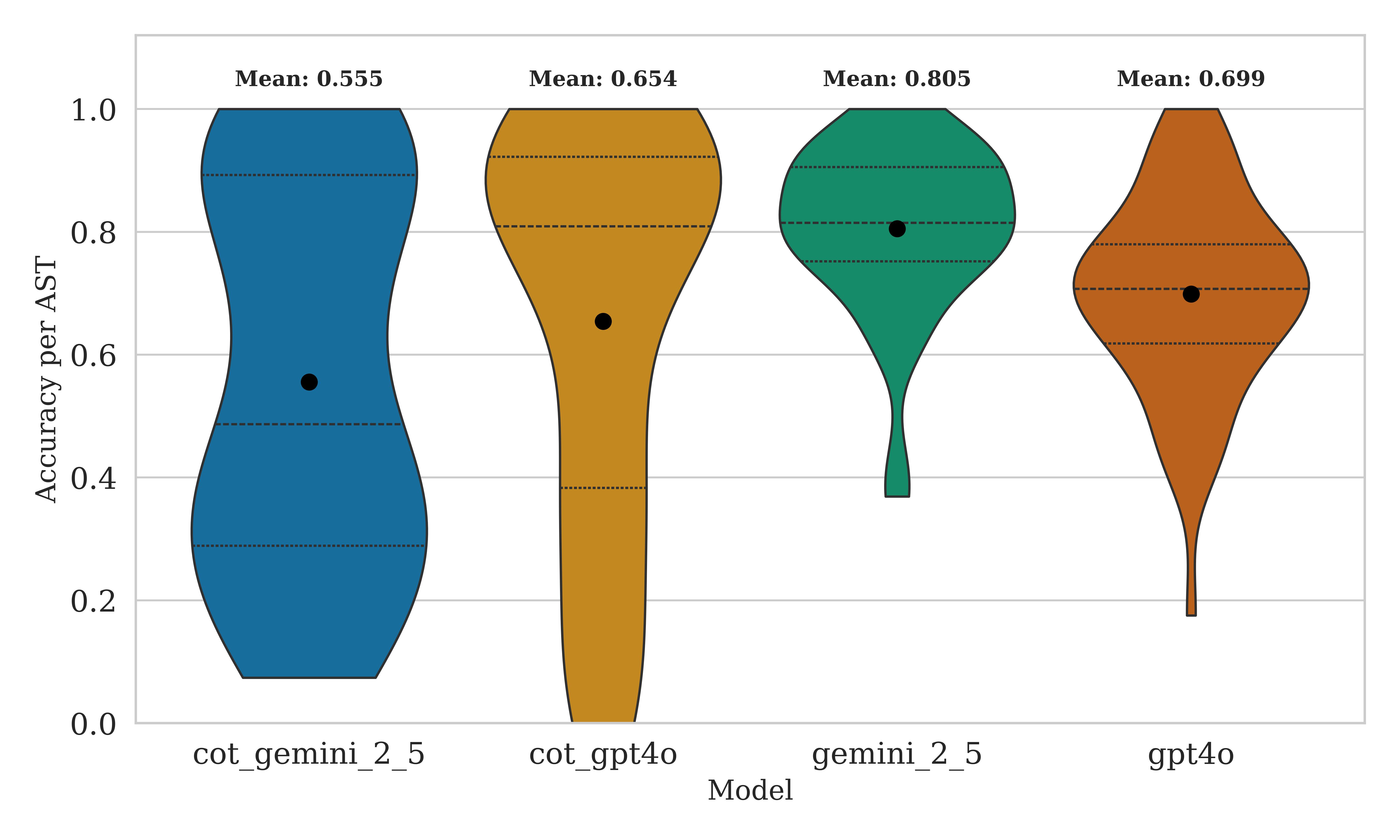}
\caption{
\textbf{Accuracy distribution across premise tags.}
Each premise tag corresponds to a distinct reasoning unit
within the VPP evaluation pipeline.
The distributions show that premise-level accuracy varies
significantly across tags, reflecting differences in the
difficulty of perception, value extraction, and relational reasoning
tasks required by different premises.
}
\label{fig:accuracy_tag_distribution}
\end{figure*}

\begin{figure*}[ht]
\centering
\includegraphics[width=0.85\linewidth]{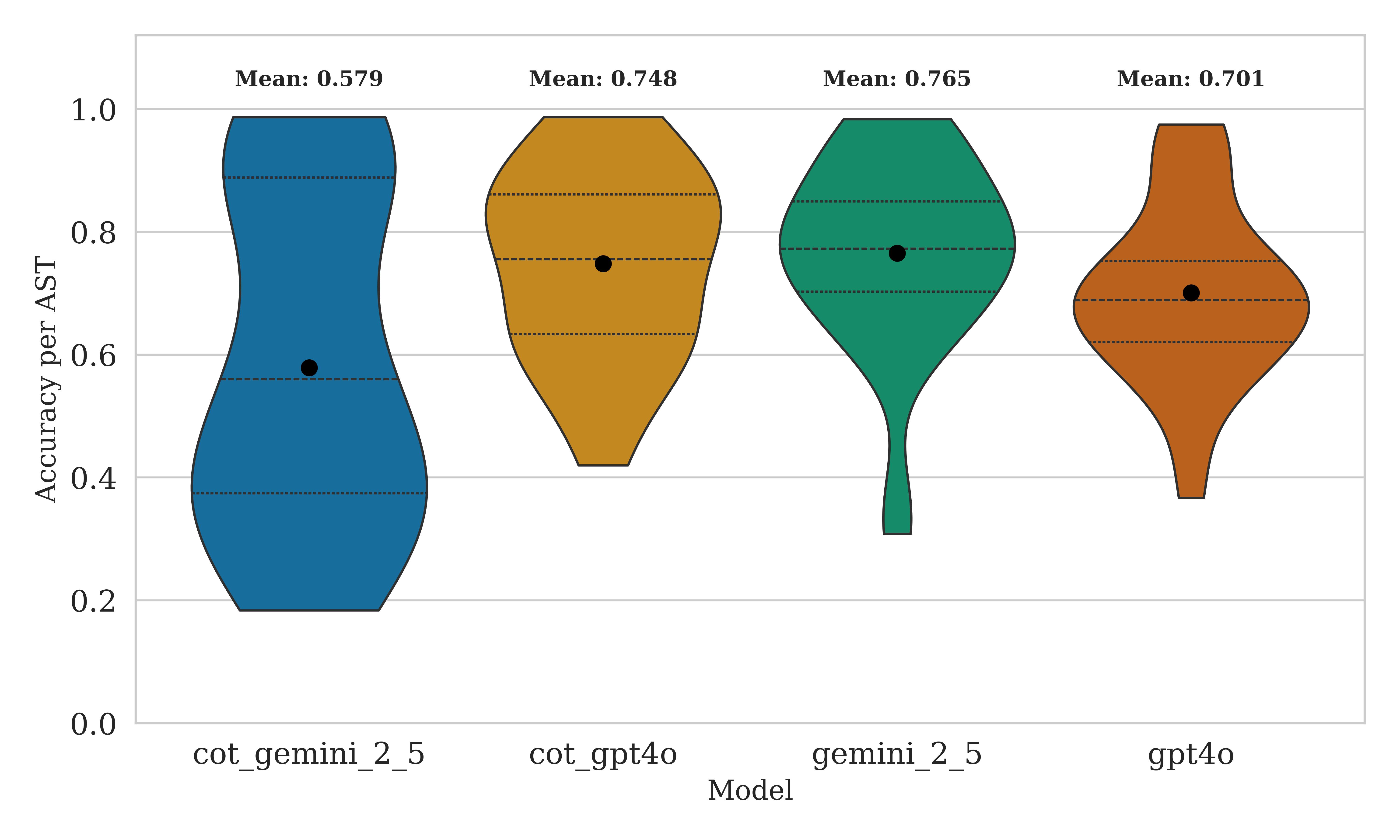}
\caption{
\textbf{Accuracy across reasoning chains (ASTs).}
Each violin summarizes model performance over complete reasoning
chains connecting structural (SP), data (DP), and relational (RP)
premises. The spread of the distributions indicates that
errors accumulate across reasoning steps: even models that perform
well on individual premises may struggle to maintain consistent
accuracy across entire chains.
}
\label{fig:accuracy_qid_distribution}
\end{figure*}

\begin{figure*}[ht]
\centering
\includegraphics[width=0.85\linewidth]{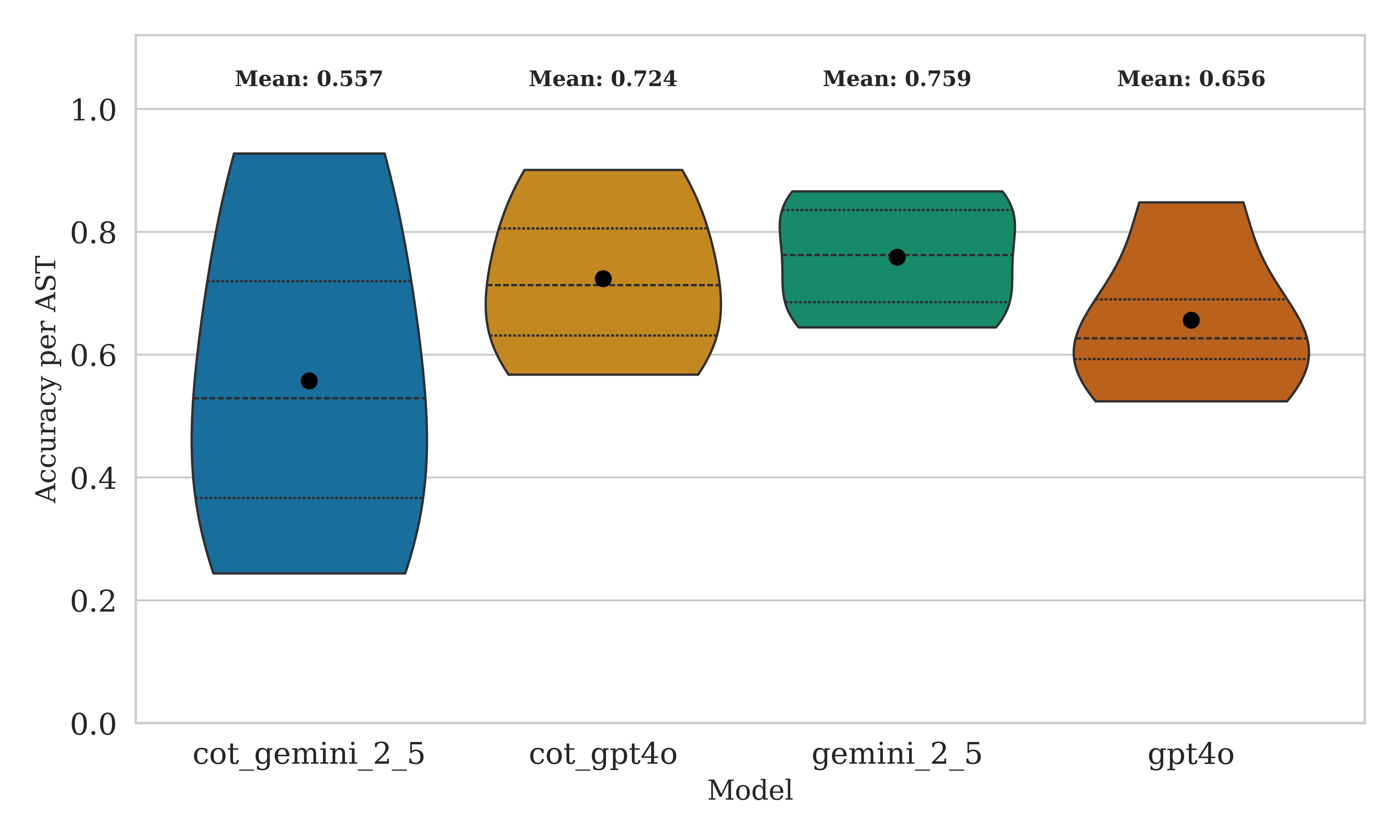}
\caption{
\textbf{Accuracy by premise source type.}
Results are grouped by the structural role of premises in the
VPP pipeline: structural perception (SP), data extraction (DP),
relational reasoning (RP), and multi-hop reasoning (MP).
Performance degrades as reasoning complexity increases,
with relational and multi-hop premises exhibiting the largest
performance variance across models.
}
\label{fig:accuracy_source_distribution}
\end{figure*}

Figures~\ref{fig:accuracy_image_distribution}--\ref{fig:accuracy_source_distribution}
visualize these distributions using violin plots. Each violin shows
the kernel density of group-level accuracy values, allowing us to
observe both central tendency and variability. The dashed lines
indicate quartiles, while the central marker denotes the mean.

This analysis reveals how reasoning difficulty is distributed across
the dataset and where errors tend to concentrate within the
compositional structure of chart question answering.

Specifically, we examine accuracy distributions across four structural
groupings:

\begin{itemize}
\item \textbf{Figure~\ref{fig:accuracy_image_distribution} (Image-level aggregation).}
Predictions are grouped by chart image, and accuracy is computed over
all premises derived from the same image. This reveals how reasoning
difficulty varies across individual charts.

\item \textbf{Figure~\ref{fig:accuracy_tag_distribution} (Premise-tag aggregation).}
Predictions are grouped by premise tag, corresponding to distinct
reasoning operations within the VPP pipeline.

\item \textbf{Figure~\ref{fig:accuracy_qid_distribution} (Reasoning-chain aggregation).}
Predictions are grouped by question identifier (AST), representing
complete reasoning chains composed of multiple premises.

\item \textbf{Figure~\ref{fig:accuracy_source_distribution} (Premise-type aggregation).}
Predictions are grouped by structural premise type
(SP, DP, RP, MP), enabling comparison between perceptual grounding,
data extraction, and higher-order relational reasoning tasks.
\end{itemize}

These plots complement the main quantitative results by revealing
where models succeed or fail within the compositional reasoning
structure of chart question answering.

\vspace{8pt}

\begin{figure*}[ht] \centering \includegraphics[width=0.85\linewidth]{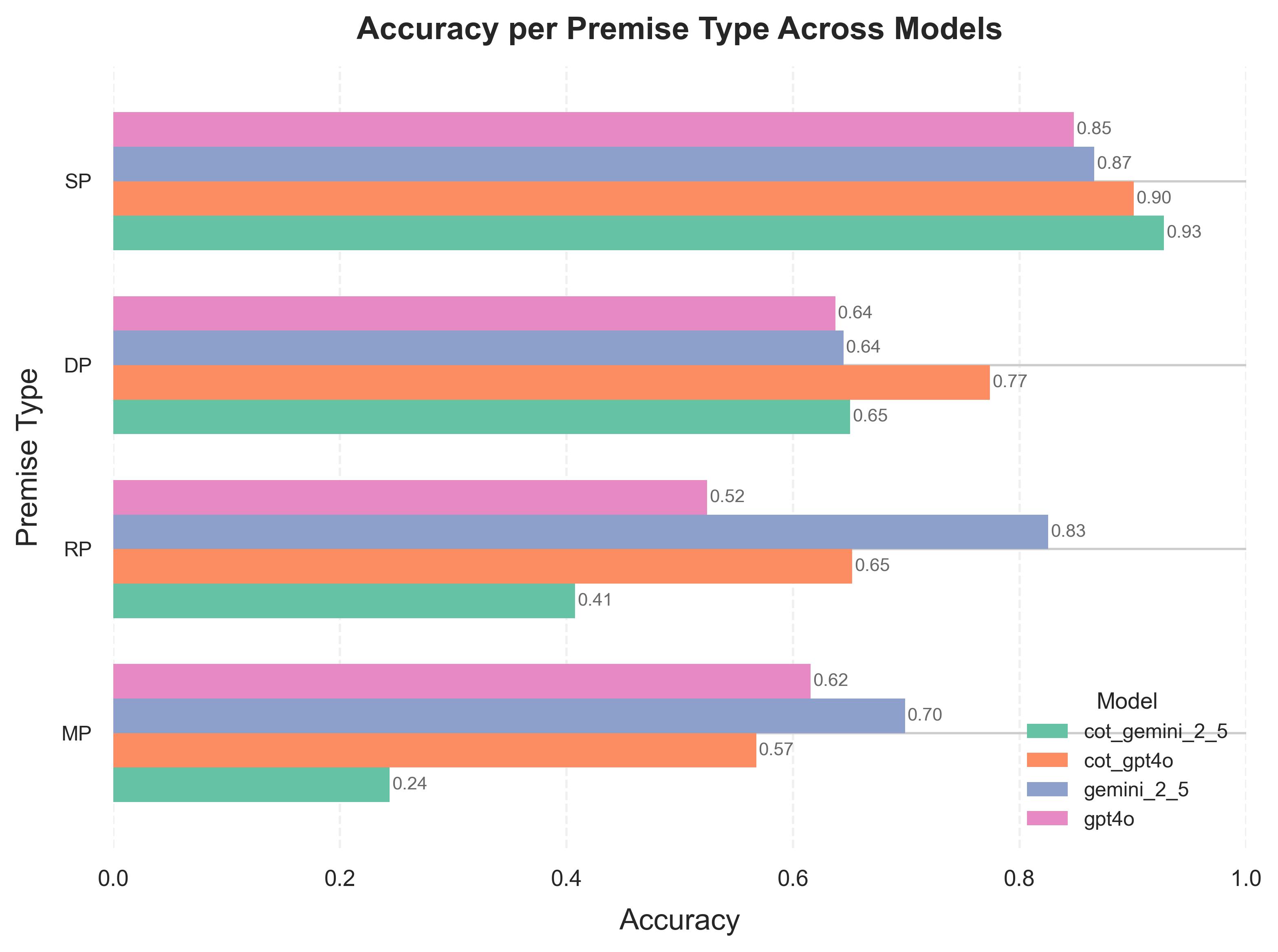} \caption{ Accuracy per premise tag across models. Tags are sorted by average accuracy, highlighting strengths and weaknesses in tag-level reasoning performance. } \label{fig:accuracy_bar_source2} \end{figure*} \begin{figure*}[ht] \centering \includegraphics[width=0.85\linewidth]{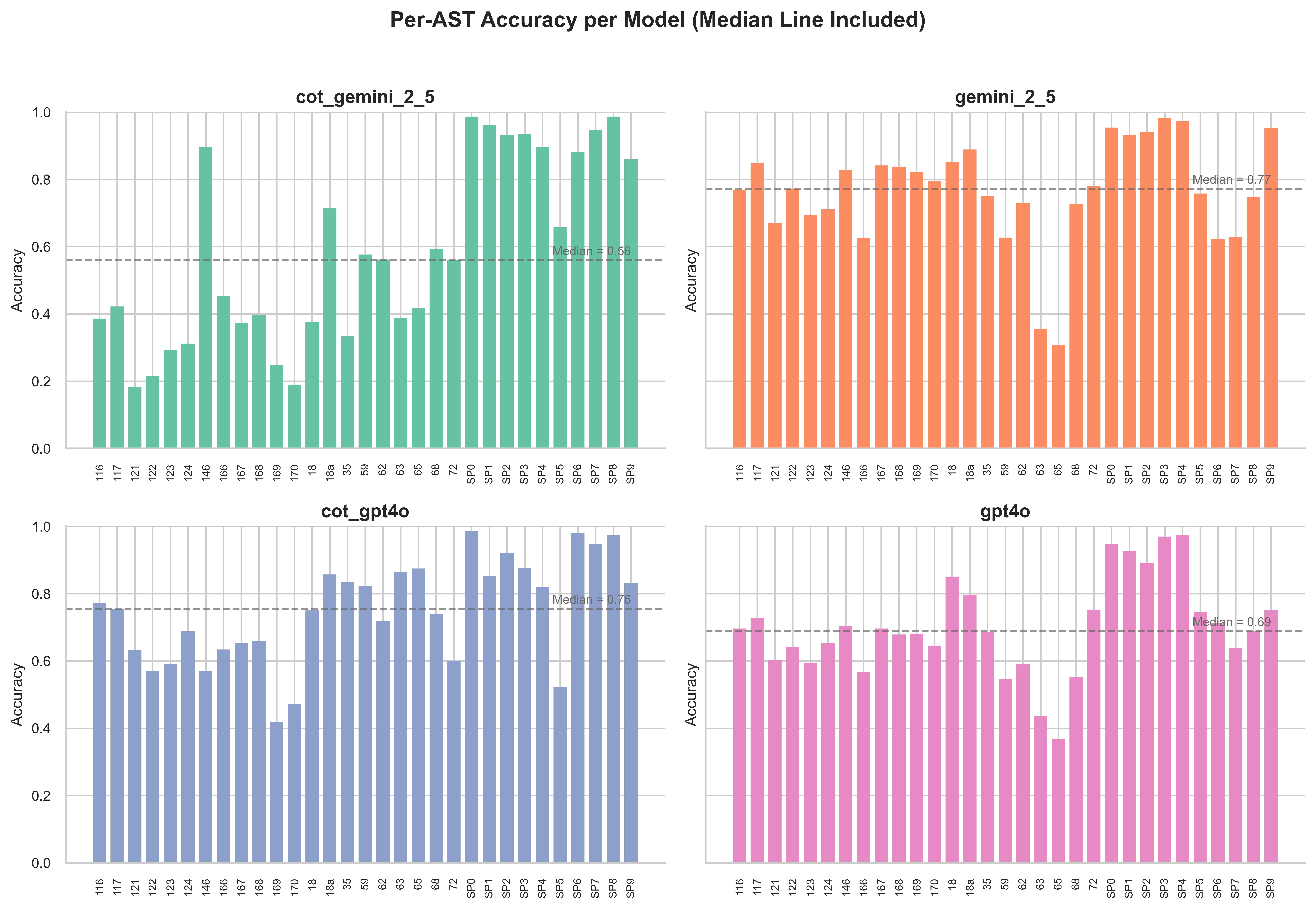} \caption{ Accuracy per AST across models. } \label{fig:accuracy_qid2} \end{figure*}

\subsection{Reasoning Behavior Across Premise Types}
\label{app:premise_analysis}

Figure~\ref{fig:accuracy_bar_source2} summarizes model accuracy
across these premise categories. Structural premises exhibit the
highest accuracy across all models, indicating that modern
multimodal systems perform relatively well on perceptual grounding
tasks such as identifying axes, legends, or chart structure.
Data premises, which require extracting specific values or entities
from the chart, show moderately lower performance but remain
manageable for most models.

In contrast, relational and multi-hop premises show substantially
lower accuracy. These premises require reasoning over multiple
entities or performing arithmetic comparisons between chart elements.
The drop in performance highlights a persistent weakness in
compositional reasoning over structured visual data.

\subsection{Reasoning Chain Variability}

While premise-level analysis reveals localized reasoning weaknesses,
full reasoning chains provide a more holistic view of model behavior.

Figure~\ref{fig:accuracy_qid2} visualizes accuracy across complete
reasoning chains (ASTs) for each model. Each AST corresponds to a
structured reasoning path connecting SP, DP, and RP premises.

The distribution reveals two key patterns. First, models exhibit
significant variability across reasoning chains, indicating that
difficulty is strongly dependent on the specific structural
configuration of premises. Second, models that perform well on
individual premises do not always maintain consistent performance
across the entire reasoning chain. Errors in earlier premises can
propagate and lead to incorrect final conclusions.

Together, these analyses highlight the value of the VPP framework
for diagnosing reasoning failures in multimodal systems.

\clearpage
\FloatBarrier

\section{Full VQA Benchmark Results}
\label{apx:vqares}

Table~\ref{tab:vqa_full_results} reports the complete VQA evaluation
across answer types, question categories, and chart types.

We compare several chart VQA models with the proposed
\textbf{RealCQA-V2} benchmark. Our evaluation focuses primarily on
smaller pre-LLM models (typically under 20M parameters), providing
a challenging setting for visual reasoning over charts.

Results show that RealCQA-V2 significantly improves performance
across most evaluation categories, with particularly strong gains
for structural reasoning tasks and complex chart types such as
scatter plots and box plots.

\begin{figure}[ht]
  \centering
    \includegraphics[width=0.9\linewidth]{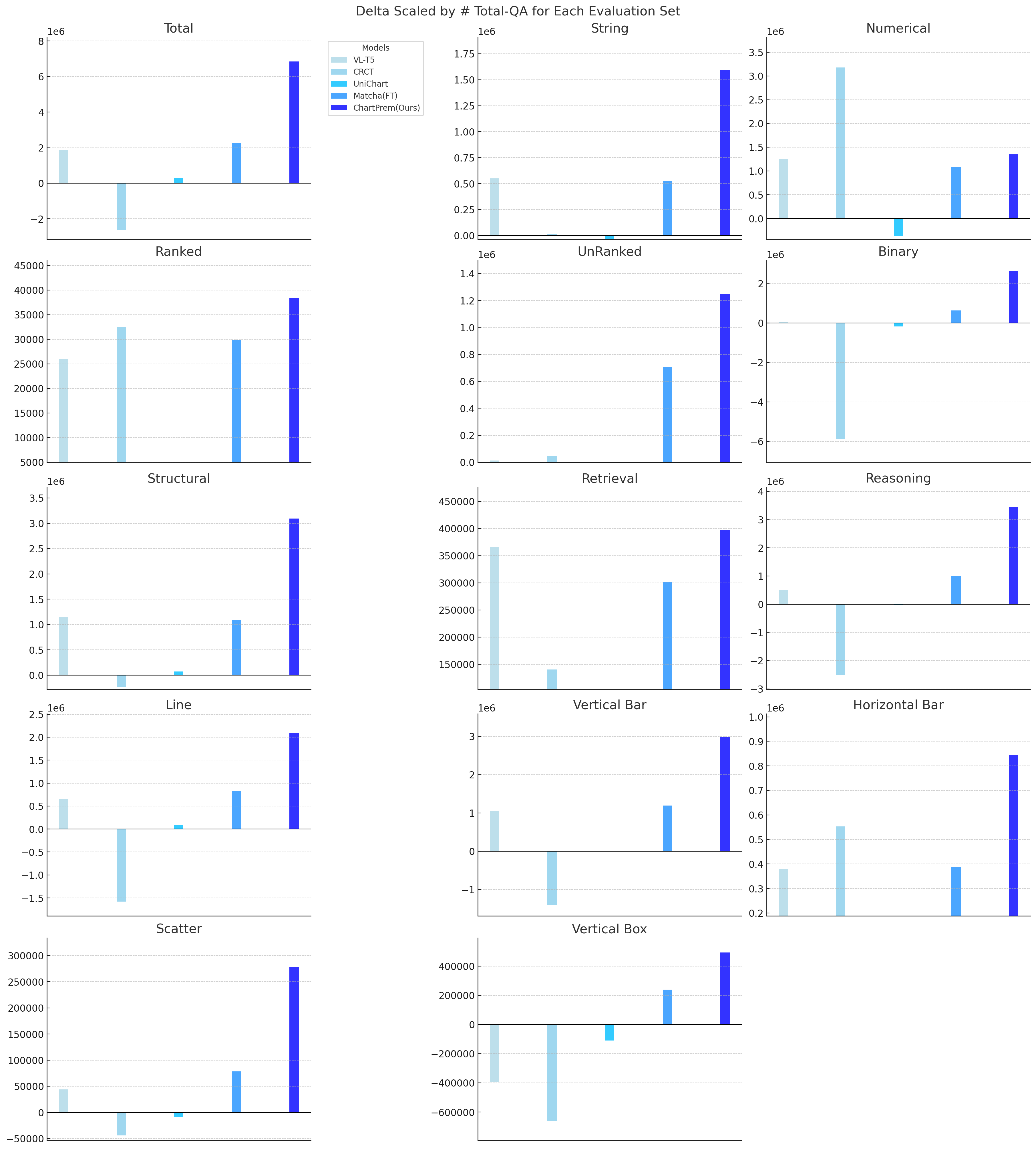}
    \caption{ Taking Zero Shot Matcha as baseline comparison over NLP-QA \scriptsize(Best viewed digital \textbf{zoom} and \textbf{color})}
    \label{fig:nlp-qa}
\end{figure}

Figure \ref{fig:nlp-qa} visualizes the results of Table \ref{tab:vqa_full_results} to showcase training on premises helps model improve on NLP-QA task.

\begin{table*}[t]
\centering
\small
\setlength{\tabcolsep}{5pt}
\renewcommand{\arraystretch}{1.15}

\begin{tabular}{l r c c c}
\toprule

 & & \multicolumn{3}{c}{Model Accuracy (\%)} \\
\cmidrule(lr){3-5}

Model & \#QA
& VL-T5
& CRCT
& UniChart \\

\midrule

\multicolumn{5}{l}{\textbf{Overall Performance}} \\

Total Accuracy
& 367,139
& 31.06
& 18.80
& 26.75 \\

\midrule

\multicolumn{5}{l}{\textbf{Answer Type Breakdown}} \\

String Answer
& 19,525
& 30.68
& 3.23
& 0.88 \\

Numerical Answer
& 115,391
& 14.87
& 31.58
& 0.83 \\

Unranked Answer
& 44,702
& 0.48
& 1.24
& 0.14 \\

Binary Answer
& 171,132
& 52.75
& 18.07
& 51.53 \\

\midrule

\multicolumn{5}{l}{\textbf{Question Type Breakdown}} \\

Structural Questions
& 48,306
& 43.52
& 14.98
& 21.40 \\

Retrieval Questions
& 8,220
& 58.77
& 31.31
& 24.72 \\

Reasoning Questions
& 310,613
& 29.37
& 19.60
& 27.64 \\

\midrule

\multicolumn{5}{l}{\textbf{Chart Type Breakdown}} \\

Line Charts
& 115,899
& 38.24
& 19.06
& 33.51 \\

Vertical Bar Charts
& 178,740
& 28.79
& 15.06
& 22.99 \\

Horizontal Bar Charts
& 46,214
& 25.42
& 29.17
& 20.58 \\

Scatter Charts
& 4,371
& 28.29
& 8.07
& 16.09 \\

Vertical Box Charts
& 21,915
& 24.06
& 11.84
& 36.93 \\

\midrule
\multicolumn{5}{c}{\textbf{Matcha(Base) trained on Different datasets}} \\
\midrule

\toprule
 Dataset& & Matcha & RealCQA & \textbf{RealCQA-V2} \\
\midrule

Total Accuracy
& 367,139
& 25.97
& 32.10
& \textbf{44.62} \\

\midrule

String Answer
& 19,525
& 2.47
& 29.50
& \textbf{83.97} \\

Numerical Answer
& 115,391
& 4.01
& 13.39
& \textbf{15.68} \\

Unranked Answer
& 44,702
& 0.20
& 16.03
& \textbf{28.11} \\

Binary Answer
& 171,132
& 52.54
& 56.19
& \textbf{67.95} \\

\midrule

Structural Questions
& 48,306
& 19.85
& 42.41
& \textbf{83.89} \\

Retrieval Questions
& 8,220
& 14.20
& 50.82
& \textbf{62.44} \\

Reasoning Questions
& 310,613
& 27.71
& 30.89
& \textbf{38.84} \\

\midrule

Line Charts
& 115,899
& 32.67
& 39.78
& \textbf{50.72} \\

Vertical Bar Charts
& 178,740
& 22.95
& 29.60
& \textbf{39.69} \\

Horizontal Bar Charts
& 46,214
& 17.19
& 25.56
& \textbf{35.45} \\

Scatter Charts
& 4,371
& 18.19
& 36.09
& \textbf{81.81} \\

Vertical Box Charts
& 21,915
& 41.99
& 52.86
& \textbf{64.52} \\

\bottomrule
\end{tabular}

\caption{
Comprehensive VQA evaluation on the RealCQA benchmark across multiple answer types,
question categories, and chart visualization types.
Underlined models are evaluated in the zero-shot setting, while others are fine-tuned.
RealCQA-V2 consistently improves performance across most evaluation categories,
particularly for structural reasoning and complex chart types.
}

\label{tab:vqa_full_results}
\end{table*}

\clearpage
\FloatBarrier

\section{Dataset Details}
\begin{table}[ht!]
\centering
  \scriptsize
  \begin{tabular}{|l|c|c|c|c|}
  \hline
    Dataset& \# Img  & Source & Task \\
    \hline
    FigureSeer \cite{Siegel2016FigureSeerPR} & 1k & ArXiv & Dense  \\
    
    ChartInfo \cite{davila2022icpr} & 28k  & PubMed Central & Dense  \\
    Real CQA \cite{ahmed2023realcqa} & 28k  &PubMed Central & QA  \\
    ChartQA \cite{masry2022chartqa} & 22k  & Pew/Statista/OWID & QA \\

    C2T \cite{obeid-hoque-2020-chart} & 82k  & Pew/Statista& Summary \\
    
    EC400k \cite{luo2021chartocr} & 400k & Excel & Dense  \\
    FQA \cite{kahou2018figureqa} & 180k  & Synthetic & QA \\
    DVQA \cite{} & 300k  & Synthetic & QA \\
    LeafQA \cite{DBLP:journals/corr/abs-1907-12861} & 200k & Synthetic & QA \\ 
    \hline

\end{tabular}
\caption{Popular Chart Datasets, Sources and Tasks}
  \label{tab:freq}
\end{table}

Synthetic datasets like FQA, DVQA, and LeafQA, have extensive scale (180k to 300k charts) and dense text annotations(components, captions, summaries, question answers), usually leverage real world tabular data-source and plot images using standard libraries like Matplotlib.  

Real-world chart datasets such as, FigureSeer, UB-PMC, and EC400k, are costly to annotate and have limited number of images/annoations available. Scientific charts, from sources like ArXiV and PMC encompass a wide range of technical and stochastic data required for academic discourse as compared to business oriented excel charts. 

While synthetic charts are easy to scale they lack fidelity with real-world chart images and under-perform with even a slight variation in the data distribution. Digital-born scientific publications, especially pose a significant challenge due to their complex visual layouts and intricate details, such as dense plot elements, noisy overlapping lines and bars, high concentration of math and special symbols etc. 

FigureSeer \cite{kahou2018figureqa},  consists of $\sim$1k densely annotated line charts from arXiv publications. RealCQA consists of $\sim$ 2Mil QA pairs based on 240 templates for $\sim$28k human annotated charts first proposed by UB-PMC from pubmed central publications. EC400k provides line, bar and pie plots from business based sources and obfuscates text in charts, making any further semantic use impossible. 
ChartQA, C2T used for chart to text tasks are  primarily based on more straight forward data from sources like Pew, Statista and Our World in Data(OWID).

We base our study on RealCQA dataset as it is the only dataset that provides a combination of challenging real-world data, scientific chart images, and dense annotations of both structural elements and textual information to ensure the creation of verifiable FOL reasoning sequences. Two requirements for creating a valid FOL are (i) a closed set of variables and (ii) a closed set of predicates. The closed set of variables includes chart components such as tick values, axis titles, legend labels, etc manually identified for the chart structure prediction task of UB-PMC. Further the QA templates used for RealCQA were handcrafted by domain experts, and generate reasoning-based questions by performing mathematical comparisons between a given subset of chart components a.k.a our variables. This ensures completeness of predicate logic. We provide exhaustive details over variables, predicates, premises, and our curated FOL sequences for each template in supplementary section. 

\paragraph{Chart Question Answers} The underlying chart images and questions are taken from RealCQA we refer reader to for exhaustive details. 

\paragraph{Chart FOL Details}
To convert a question about a chart with binary answers to first-order logic (FOL), we need to represent the relevant structural elements of the chart and the relationships between them.

\paragraph{Chart Variables}
\begin{itemize}
    \item \small{$X$-axis title (\textit{Xtitle})}
    \item \small{$i$-th $X$-axis tick marks (\textit{Xi})}
    \item \small{Closed range of values of $X$-ticks [$x_0, x_n$]}
    \item \small{$Y$-axis title (\textit{Ytitle})}
    \item \small{$j$-th $Y$-axis tick mark (\textit{Yj})}
    \item \small{Closed range of values of $Y$-ticks [$y_0, y_m$]}
    \item \small{Legend labels (\textit{Legendlabel})}
    \item \small{$k$-th legend label (\textit{Lk})}
    \item \small{Closed range of values of legends, i.e., data series names [$l_0, l_h$]}
    \item \small{$i+1/j+1$ represents successive $i$-th/$j$-th value of the respective variable}
\end{itemize}

\paragraph{Chart Predicates}
\begin{enumerate}
    \item $\exists(\{Xi_0, \ldots, Xin\}, Xtitle)$: for all $X$ ticks of $Xtitle$ in $C$, there exist a given set of $X$ tick values, where $Xi$ denotes the $i$-th $X$ tick value.
    
    \item $\exists(\{Yj_0, \ldots, Yjm\}, Ytitle)$: for all $Y$ ticks of $Ytitle$ in $C$, there exist a given set of $Y$ tick values, where $Yj$ denotes the $j$-th $Y$ tick value.
    
    \item $\exists(\{Lh_0, \ldots, Lhk\}, Legendlabel)$: for all labels in $Legendlabel$ in $C$, there exists a set of given labels, where $Lk$ denotes the $k$-th label.
    
    \item $Value\_At(\{(Xi_0, Yj_0), \ldots, (Xin, Yjm)\}, Ytitle)$: the value of $Ytitle$ at each data point $(Xi_0, Yj_0), \ldots, (Xin, Yjm)$ exists in $C$.
    
    \item {\scriptsize $Value\_At(\{(Xi_0, Yj_0, Lk_0), \ldots, (Xin, Yjm, Lkh)\}, Ytitle, Legend)$}
    
    the value of $Ytitle$ for the $k$-th given $Legend$, $Lkh$, at each data point in $\{(Xi_0, Yj_0, Lk_0), \ldots\}$ exists in $C$.
    
    \item $Max\_Value((Xi, Yj), Ytitle)$: the data point represented by $(Xi, Yj)$ is the maximum value of $Ytitle$ across all data points in $C$.
    
    \item $Max\_Value((Xi, Yj, Lk), Ytitle, Legendlabel)$: the data point represented by $(Xi, Yj)$ is the maximum value of $Ytitle$ for the given $Legendlabel$ across all data points in $C$.
\end{enumerate}

These conditions ensure that the chart $C$ has valid and complete data, as well as allowing for comparison of data across different data points and legends. 

\paragraph{Chart Premises Creation} :
Our process of creating premises relies on deconstructing the binary reasoning questions of the RQA dataset. This is done inspired from a bottom up method of building up from first principles how a human being reads a chart. This involves certain common steps of identifying the chart structure and a unique premise per original reasoning question. Mathematical reasoning questions are deconstructed to base arithmetic steps to calculate the particular value. The premises are created as individual statement, conclusion pair per the question templates. We use a T5 transformer to further create 2-3 paraphrases of each for vocabulary diversity. Then for each question the premise templates are populated with chart specific values and generate both positive and negative cases.

\begin{figure*}[!ht]
  \centering
  \begin{subfigure}[b]{0.45\textwidth}
    \centering
    \includegraphics[width=\linewidth]{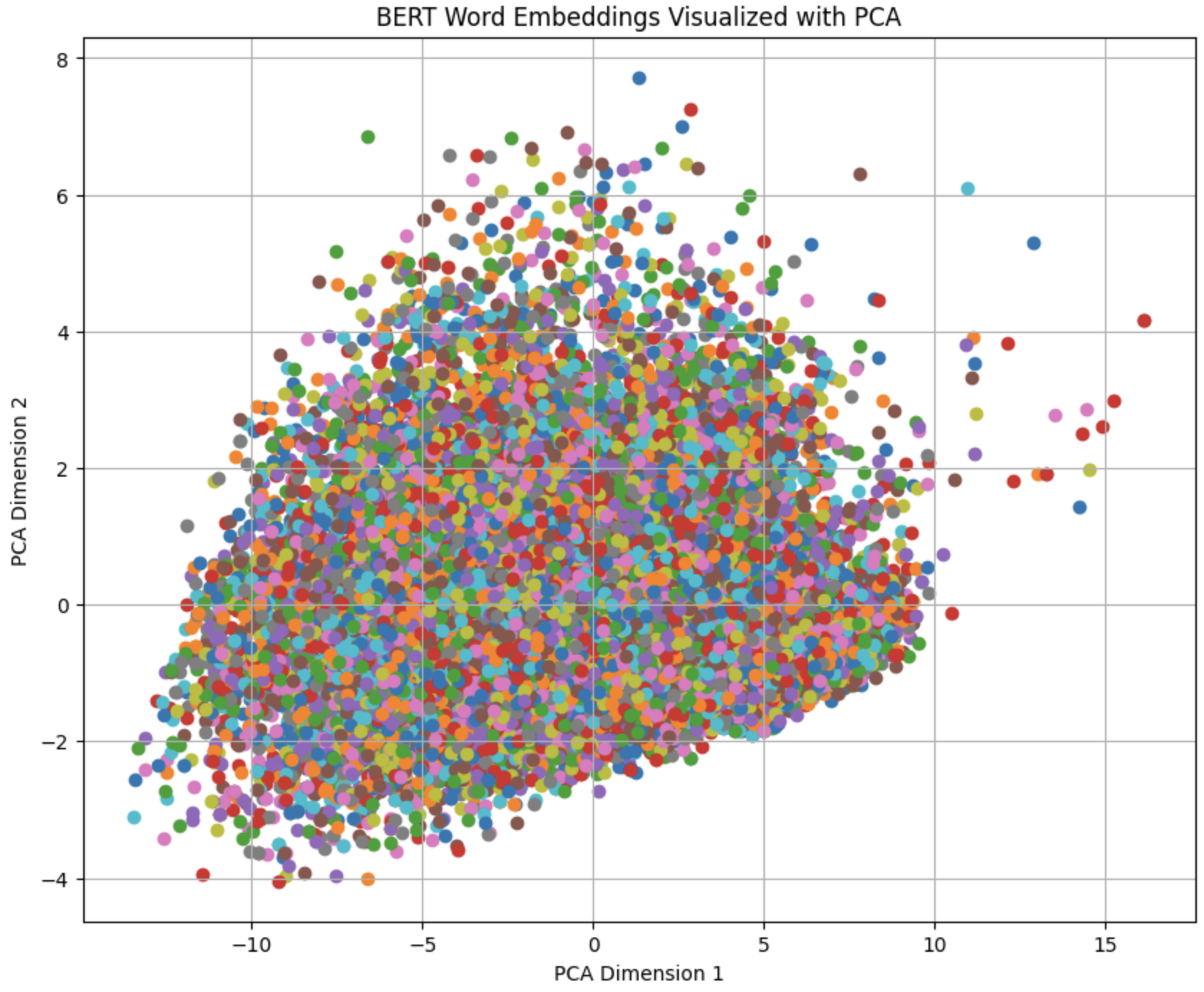}
    \caption{PCA }
    \label{fig:PCA}
  \end{subfigure}%
  \begin{subfigure}[b]{0.45\textwidth}
    \centering
    \includegraphics[width=\linewidth]{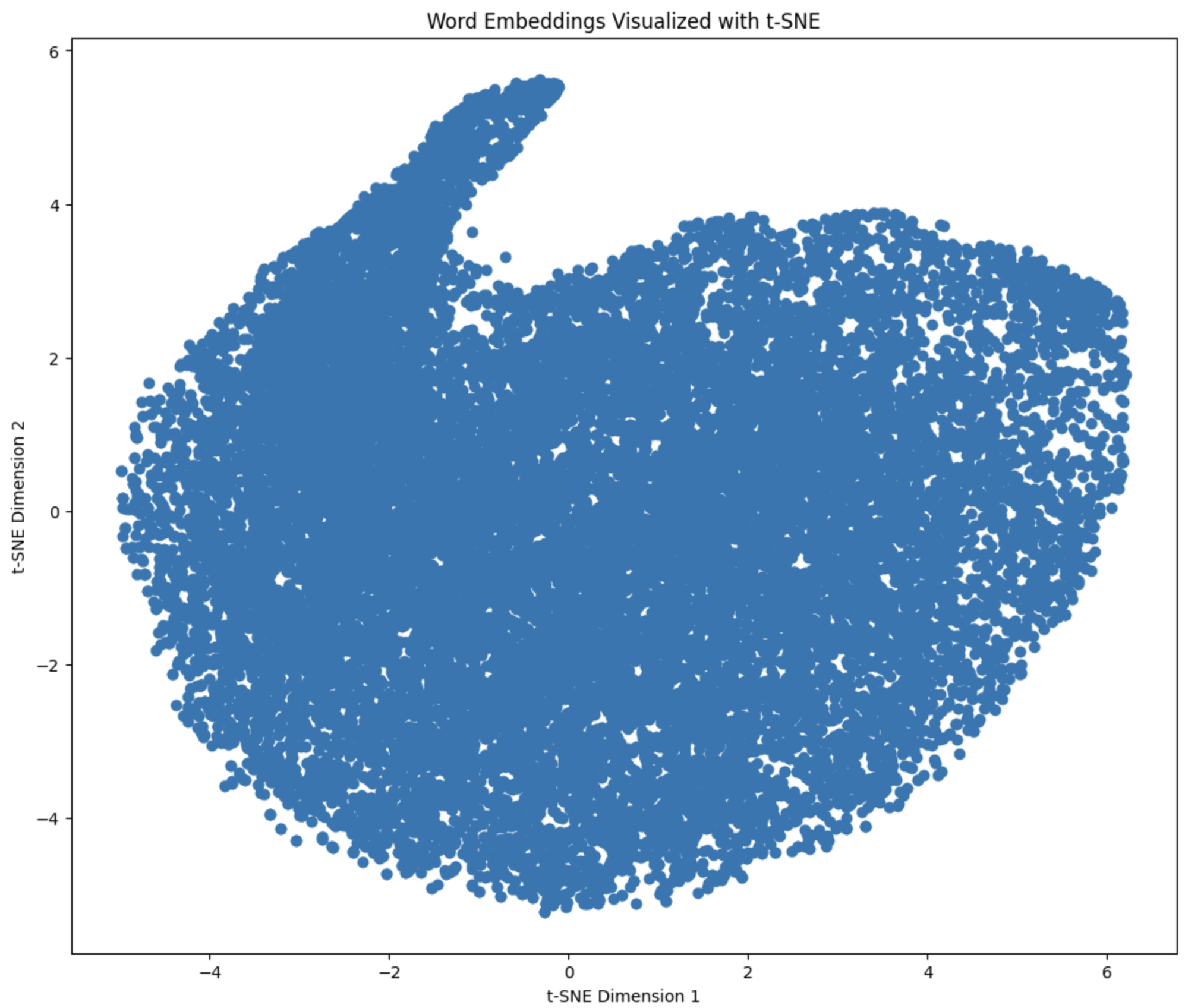}
    \caption{tSNE }
    \label{fig:tSNE}
  \end{subfigure}

  \caption{Distribution of NLP Premises \scriptsize{(Best viewed digital, \textbf{zoom} and \textbf{color})}}
  \label{fig:combined}
\end{figure*}

The PCA and t-SNE plots of pretrained-BERT embeddings as shown in Figures \ref{fig:PCA}, and \ref{fig:tSNE} for 25,364 unique words indicate a highly diverse vocabulary, with a broad and evenly distributed semantic space and no apparent clustering, highlighting rich semantic coverage. In contrast to the focused and less diverse vocabularies in popular VQA tasks, which show distinct clusters around common categories and actions, this corpus encompasses a wider range of topics and semantics. The extensive diversity and rich semantic distribution suggest a more nuanced and challenging dataset.

\subsection{Dense Chart Parsing}
The task refers to locating and recognizing each individuval component of a chart.

\subsection{ The Chart Infographics Challenge}
ChartInfo Challenge \cite{davila2022icpr} and its previous iterations span about half a decade worth of research. Proposed initially in 2019, as an overarching task to extract tabular data from charts the first iteration saw a very large synthetic dataset and a small real world dataset used for eval only. This was quickly scaled up in the later iterations of the challenge. The charts are taken from digitally born pdf of scientific publications from the open access subset of PubMed Central. 
The challenge consisted of 7 tasks aimed at (i) Chart Type Classification (ii/iii) Chart Text location,recognition and role (iv) Axis and Tick Location (v) Legend location and Mapping (vi) Tabular Data location and Extraction (vii) End-to-End Data extraction. The real world charts were painstakingly manually annotated for each of these tasks and forms the foundation of our work today.

\subsection{Results}
In Table \ref{tab:res1_dense}, we compare our model's performance on dense chart parsing tasks by using structured premises for structure prediction, such as querying 'What is the type of chart?' or 'What is the title of the dependent axis?' instead of conventional image classification. We compare with the  results of the corresponding tasks from the previous challenge\cite{davila2022icpr}, and in the column `Direct Prediction' we report results as reported in the challenge report by task specific vision models. The next column is the Matcha-base model in a zero shot setting, and the next column `Matcha-FT' is the same model when trained only on the RealCQA QA pairs. The next column, `ChartPrem' is our proposed model trained further on the SP's we created. We provide exhaustive list of our queries in the appendix. These queries are complementary to the underlying binary SP's that we created for the VPP task and have text answers. Thus, training on SP's improves performance on previous dense chart parsing tasks as compared to direct pixel level predictions. While zero-shot Matcha is only able to generalize sufficiently to three chart properties categorical labels, logarithmic axis and presence of legend, on fine-tuning the performance improves but the vision based task specific models still outperform. Only on further training with the SP's we see considerable improvement. The worst performance is for dependent axis title which can include multiple complex math symbols used in scientific charts which the model might not have seen as much due to the text heavy pre-training datasets. The second worst performance is for categorical x-tick labels and this is due to the complex grouped and stacked bar charts which might again have math-symbol intensive labels and at times are quite cluttered having 40-50 tilted labels on a single chart.

\begin{table}[t]
\small
\centering
\begin{tabular}{lcccc}
\toprule
\textbf{Chart Component Task} & \textbf{Direct} & \underline{\textbf{Matcha}} & \textbf{RealCQA} & \textbf{RealCQA-V2} \\
\midrule

\textbf{Chart Structure} \\

Chart Type {\small(F1)} & 94.63 & 35.21 & 80.96 & \textbf{98.72} \\
Dependent (Y) Axis Title {\small(Text)} & 75.62 & 21.85 & 47.82 & \textbf{77.80} \\
Independent (X) Axis Title {\small(Text)} & 85.62 & 42.30 & 79.34 & \textbf{92.38} \\

\midrule

\textbf{Axis Value Extraction} \\

Y-Min Value {\small(Abs)} & 73.42 & 34.61 & 70.10 & \textbf{97.59} \\
Y-Max Value {\small(Abs)} & 62.22 & 24.74 & 54.63 & \textbf{96.50} \\
X-Min Value {\small(Abs)} & 73.42 & 24.61 & 62.77 & \textbf{96.94} \\
X-Max Value {\small(Abs)} & 62.22 & 34.74 & 57.02 & \textbf{95.31} \\

\midrule

\textbf{Categorical Structure} \\

Categorical X-Tick Labels {\small(1:1)} & 68.83 & 33.78 & 58.76 & \textbf{79.74} \\

\midrule

\textbf{Chart Metadata} \\

Is Categorical {\small(Binary)} & -- & 82.97 & 91.48 & \textbf{100.00} \\
Is Logarithmic {\small(Binary)} & -- & 84.65 & 93.24 & \textbf{100.00} \\
Legend Present {\small(Binary)} & -- & 77.32 & 90.66 & \textbf{100.00} \\
Number of Data Series {\small(Abs)} & -- & 38.43 & 56.57 & \textbf{82.61} \\
Legend Name {\small(1:1)} & 82.93 & 47.92 & 73.86 & \textbf{91.27} \\

\bottomrule
\end{tabular}

\caption{
Dense chart parsing results on ChartInfo Challenge '22.
\underline{Underlined} models denote zero-shot performance,
while the remaining models are fine-tuned.
}
\label{tab:res1_dense}
\end{table}

\FloatBarrier
\subsection{Example Structure Premises}

\begin{figure}[H]
\centering
\includegraphics[width=0.55\linewidth]{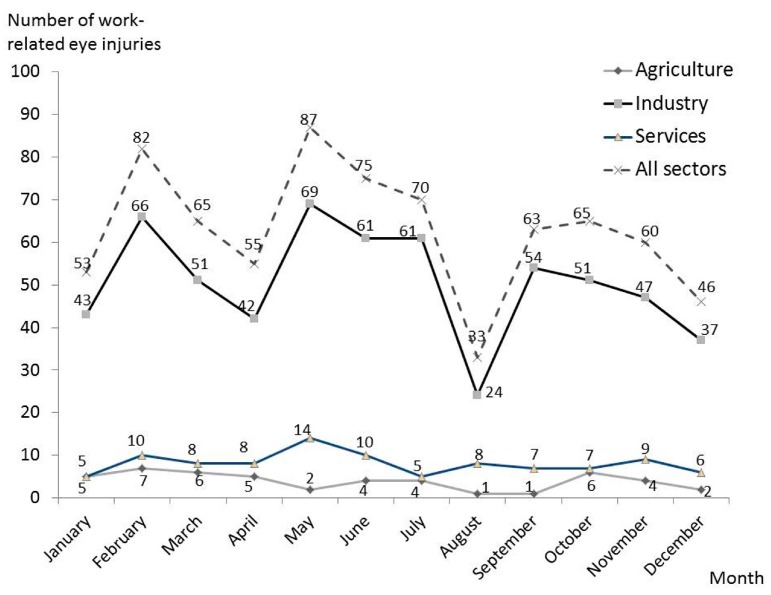}

\vspace{6pt}

\begin{scriptsize}
\textbf{SP0 (Chart Type)}  
True: The chart type is \textit{line}.  
False: heatmap; bar; scatter.

\textbf{SP1 (Dependent Axis Title)}  
True: ``Number of work-related eye injuries''.  
False: ``8''; ``July''; ``March''.

\textbf{SP2 (Independent Axis Title)}  
True: ``Month''.  
False: ``30''; ``8''; ``2''.

\textbf{SP3 (Dependent Axis Range)}  
True: Range $0$–$100$ (Number of work-related eye injuries).  
False: $0$–Services in $30$; $5$–$9$ in May; $65$–Industry in $70$.

\textbf{SP5 (Categorical X-axis)}  
True: Labels = [January, February, March, April, May, June, July, August, September, October, November, December].  
False: January; 82.

\textbf{SP6 (X-axis Tick Marks)}  
True: Tick marks correspond to Month values.  
False: Services; 61; 9.

\textbf{SP7 (Y-axis Tick Marks)}  
True: Tick marks correspond to Number of work-related eye injuries values.  
False: 82; 2; 90.

\textbf{SP8 (Legend Presence)}  
True: Legend differentiates the four data series.  
False: 65 series; January series; 1 series.

\textbf{SP9 (Legend Labels)}  
True: [Agriculture, Industry, Services, All sectors].  
False: Month; 9; 33.
\end{scriptsize}

\caption{Example structure premises generated for chart PMC5486290.  
Each premise contains a ground-truth statement and multiple
contrastive negatives used for premise-level evaluation.}
\label{fig:structure_premise_example}
\end{figure}

\subsection{Premise Templates}

The VPP framework defines a set of structured premise templates used to
construct reasoning chains over chart images. Premises are grouped into
four categories: \textbf{Structural Premises (SP)}, \textbf{Data Premises (DP)},
\textbf{Reasoning Premises (RP)}, and \textbf{Mathematical Premises (MP)}.

\paragraph{Structural Premises (SP)}
Structural premises describe the chart layout and metadata and are
generated once per chart image. Each premise is evaluated as a
\textit{True/False} statement.

\begin{tcolorbox}[breakable,colback=white,colframe=black!20]
SP0: The type of chart is \{chart\_type\}. \\
SP1: The dependent axis is labeled as \{y\_title\}. \\
SP2: The independent axis is labeled as \{x\_title\}. \\
SP3: The dependent axis ranges from \{ymin\} to \{ymax\} in \{y\_title\}. \\
SP4: The independent axis ranges from \{xmin\} to \{xmax\} in \{x\_title\}. \\
SP5: The independent axis is categorical with labels \{x\_ticks\}. \\
SP6: Tick marks corresponding to \{x\_title\} values exist on the x-axis. \\
SP7: Tick marks corresponding to \{y\_title\} values exist on the y-axis. \\
SP8: The chart contains a legend differentiating \{number\_of\_ds\} data series. \\
SP9: Legend entries correspond to unique visual representations with labels
\{legend\_labels\}.
\end{tcolorbox}

The following query templates are used for dense chart parsing tasks.

\begin{tcolorbox}[breakable,colback=white,colframe=black!20]
SP0: What is the chart type? \\
SP1: What is the label of the dependent axis? \\
SP2: What is the label of the independent axis? \\
SP3: What is the range and title of the dependent axis? \\
SP4: What is the range and title of the independent axis? \\
SP5: Is the independent axis categorical? What are the tick labels? \\
SP6: Are tick marks present for values on the independent axis? \\
SP7: Are tick marks present for values on the dependent axis? \\
SP8: Is there a legend in the chart? How many data series exist? \\
SP9: What are the legend labels for each data series?
\end{tcolorbox}

\paragraph{Data Premises (DP)}
Data premises describe factual information extracted from chart elements
such as plot values, legends, and statistical markers.

\begin{tcolorbox}[breakable,colback=white,colframe=black!20]
DP: Value exists at coordinate (\{i\}) on axis \{y\_title\}. \\
DP: Value exists at coordinate (\{i\}) for legend \{legend\}. \\
DP: Maximum value of \{y\_title\} occurs at (\{i\},\{j\}). \\
DP: The chart contains \{ln\_cnt\} lines. \\
DP: The chart contains \{leg\_cnt\} legend entries. \\
DP: The chart contains \{Ml\_lb\} mark labels.
\end{tcolorbox}

Additional templates capture statistical properties for box plots,
including quartiles, medians, and extrema.

\begin{tcolorbox}[breakable,colback=white,colframe=black!20]
DP: Lower quartile whisker exists at position \{i\}. \\
DP: Upper quartile whisker exists at position \{i\}. \\
DP: Median line exists at position \{i\}. \\
DP: Maximum whisker exists at position \{i\}. \\
DP: Minimum whisker exists at position \{i\}. \\
DP: Lower quartile value at \{x\} equals \{val\}. \\
DP: Upper quartile value at \{x\} equals \{val\}. \\
DP: Median value at \{x\} equals \{val\}.
\end{tcolorbox}

\paragraph{Reasoning Premises (RP)}
Reasoning premises express relational comparisons between values,
enabling logical inference over chart elements.

\begin{tcolorbox}[breakable,colback=white,colframe=black!20]
RP: Value of \{y\_title\} at x-tick \{i\} is less than that at \{j\}. \\
RP: Difference of values at ticks \{i\} and \{j\} is greater than zero. \\
RP: Combined values at ticks \{xi\} and \{xj\} exceed value at (\{i\},\{j\}). \\
RP: Interquartile ranges at ticks \{i\} and \{j\} are equal. \\
RP: Median at tick \{i\} is less than median at tick \{j\}. \\
RP: Upper quartile at tick \{i\} is less than that at tick \{j\}. \\
RP: Lower quartile at tick \{i\} is less than that at tick \{j\}. \\
RP: Maximum value at tick \{i\} is less than that at tick \{j\}. \\
RP: Minimum value at tick \{i\} is less than that at tick \{j\}.
\end{tcolorbox}

Additional reasoning templates capture monotonic trends and structural
relationships.

\begin{tcolorbox}[breakable,colback=white,colframe=black!20]
RP: Values monotonically increase along the x-axis. \\
RP: Values monotonically decrease along the x-axis. \\
RP: Number of lines equals number of legends. \\
RP: Number of lines equals number of mark labels.
\end{tcolorbox}

\paragraph{Mathematical Premises (MP)}
Mathematical premises capture statistical calculations used in reasoning
tasks.

\begin{tcolorbox}[breakable,colback=white,colframe=black!20]
MP: The median of set \{S\} is \{M\} when values are ordered. \\
MP: The mean of variable X is computed correctly. \\
MP: The mean of variable Y is computed correctly. \\
MP: Deviations from the mean are computed for X and Y. \\
MP: The Pearson correlation coefficient is calculated correctly.
\end{tcolorbox}

\subsection{Dataset Statistics}
\label{sec:dataset_statistics}


\begin{table}[H]
\centering
\small
\caption{Distribution of structural premise (SP) templates used in the dataset.}
\label{tab:sp_stats}
\begin{tabular}{lrr}
\toprule
SP Template & Count & Description \\
\midrule
SP0 & 113,064 & Chart title correctness \\
SP1 & 23,720  & Legend presence \\
SP2 & 17,868  & X-axis title verification \\
SP3 & 18,872  & Y-axis title verification \\
SP4 & 13,044  & X-axis tick range validation \\
SP5 & 9,291   & Y-axis tick range validation \\
SP6 & 17,868  & Unit association correctness \\
SP7 & 23,720  & Color mapping verification \\
SP8 & 14,848  & Style assignment consistency \\
SP9 & 14,848  & Legend disambiguation \\
\midrule
\textbf{Total} & \textbf{267,143} & -- \\
\bottomrule
\end{tabular}
\end{table}


\begin{table}[H]
\centering
\small
\caption{Distribution statistics computed per chart image.}
\label{tab:image_level_stats}
\begin{tabular}{lrrrr}
\toprule
Metric & Mean & Median & Max & Skew \\
\midrule
Questions & 101 & 15 & 6,579 & 9.61 \\
Premises  & 1,239 & 216 & 103,785 & 10.29 \\
Data Premises (DP) & 831 & 96 & 73,440 & 10.45 \\
Reasoning Premises (RP) & 381 & 52 & 26,100 & 9.54 \\
Math Premises (MP) & 26.7 & 3 & 10,608 & 40.82 \\
Reasoning Programs (AST IDs) & 2.93 & 3 & 6 & 0.53 \\
\bottomrule
\end{tabular}
\end{table}


\begin{table}[H]
\centering
\small
\caption{Distribution of reasoning chains by abstract syntax template (AST).}
\label{tab:qid_level_stats}
\begin{tabular}{lrrrrr}
\toprule
AST$_{id}$ & Questions & Total Premises & DP & RP & Avg Premises \\
\midrule
68 & 172,869 & 2,074,428 & 1,382,952 & 691,476 & 12 \\
62 & 138,511 & 1,662,132 & 1,108,088 & 554,044 & 12 \\
59 & 32,786 & 393,432 & 262,288 & 131,144 & 12 \\
166--170 & 6,999 & 132,981 & 111,984 & 20,997 & 19 \\
117 & 4,914 & 39,312 & 19,656 & 19,656 & 8 \\
\bottomrule
\end{tabular}
\end{table}


\begin{table}[H]
\centering
\small
\caption{Reasoning statistics aggregated by question template (QID).}
\label{tab:qid_stats}
\begin{tabular}{lrrrr}
\toprule
Metric & Mean & Median & Max & Skew \\
\midrule
Questions & 19,624 & 4,914 & 172,869 & 2.93 \\
Premises Total & 240,112 & 39,312 & 2,074,428 & 2.93 \\
DP & 161,054 & 19,656 & 1,382,952 & 2.91 \\
MP & 5,180 & 0 & 55,992 & 3.50 \\
RP & 73,877 & 7,872 & 691,476 & 2.94 \\
Avg Chain Length & 17.86 & 12 & 97.1 & 4.10 \\
\bottomrule
\end{tabular}
\end{table}


\begin{table}[H]
\centering
\small
\caption{Dataset split statistics for training and evaluation.}
\label{tab:split_stats}
\begin{tabular}{lcc}
\toprule
Metric & Train & Test \\
\midrule
Premises & 5.21M & 51K \\
QA Chains & 408K & 4K \\
Chart Images (PMC IDs) & 15K & 4K \\
Complete Chains & 408K & 2,197 \\
Split Chains & -- & 1,865 \\
True / False Labels & 1.33M / 3.88M & 13K / 38K \\
\bottomrule
\end{tabular}
\end{table}



\begin{table}[h]
\centering
\small
\caption{Distribution of premise types across splits.}
\label{tab:premise_types}
\begin{tabular}{lrr}
\toprule
Premise Type & Train & Test \\
\midrule
Structural (SP) & 249,254 & 14,351 \\
Data (DP) & 3,330,534 & 19,757 \\
Relational (RP) & 1,520,817 & 15,507 \\
Mathematical (MP) & 106,419 & 1,392 \\
\bottomrule
\end{tabular}
\end{table}

\begin{table}[h]
\centering
\small
\caption{Dataset metadata for the released version.}
\label{tab:dataset_metadata}
\begin{tabular}{ll}
\toprule
Property & Value \\
\midrule
Dataset Name & RealCQA-V2 (VPP Benchmark) \\
Version & v0.1 \\
Release Date & Oct 2025 \\
Total Premises & 5.26M \\
Train Size & 5,207,024 \\
Test Size & 51,007 \\
Charts (PMC IDs) & 19,137 \\
Premise Types & SP, DP, RP, MP \\
False:True Ratio & 3:1 \\
\bottomrule
\end{tabular}
\end{table}

The dataset contains over 5.26M premise verification instances
derived from more than 19K chart images. Premises are divided
into four categories: structural (SP), data (DP), relational (RP),
and mathematical (MP). The benchmark follows a controlled
3:1 False–True ratio. The train/test split is constructed at the
reasoning-chain level to minimize leakage: 99.01\% of chains
appear exclusively in the training split, while 0.53\% appear
only in the test split.
\subsection{Dataset Release and Availability}

The dataset introduced in this work will be publicly released to support
reproducibility and further research on visual reasoning over chart data.
All dataset resources, including images, annotations, evaluation scripts,
and documentation, will be made available through the project website:

\begin{center}
\url{https://cse-ai-lab.github.io/VPP/}
\end{center}

The release includes the full dataset used in our experiments together with
structured premise annotations, reasoning chains, and evaluation protocols
for the Visual Premise Proving (VPP) benchmark. We provide standardized
train, validation, and test splits along with utilities for loading the
dataset and reproducing the experiments reported in the paper.

In addition to the raw annotations, the repository includes scripts for
dataset preprocessing, premise generation, and evaluation metrics used in
the benchmark. These resources are intended to facilitate consistent
comparison across methods and to encourage the development of models that
perform structured reasoning over chart images.

All documentation necessary to reproduce the experiments and extend the
dataset is provided on the project website. The release will also include
guidelines for submitting results to the benchmark and for contributing
additional annotations or extensions to the dataset.

We hope that the public availability of the VPP benchmark will support
future work on multimodal reasoning, chart understanding, and structured
visual question answering.

\subsection{Annotation Statistics and Reviewer Agreement}
\label{app:annotation_stats}

\begin{table}[t]
\centering
\small
\begin{tabular}{l c}
\toprule
\textbf{Annotation Metric} & \textbf{Value} \\
\midrule
Number of annotators & 7 \\
Number of reviewers & 4 \\
Total premise templates authored & 200 \\
Total reasoning chains defined & 412610 \\
Chains audited (random sample) & 5--10\% \\
Audit size (chains) & 40k \\
Initial reviewer agreement & 78.96\% \\
Final agreement after adjudication & 100\% \\
Templates revised during review & 12\% \\
\bottomrule
\end{tabular}
\caption{
Summary of annotation workforce and review statistics for the VPP
premise template library and reasoning chains.
}
\label{tab:annotation_stats}
\end{table}

To provide transparency on the annotation process described in main paper, we report summary statistics of the annotation workforce,
review coverage, and reviewer agreement.

All premise templates and reasoning chains were initially authored by
graduate annotators and subsequently reviewed through a two-stage
verification process. The lead reviewer performed template-level
validation, while a review panel audited randomly sampled reasoning
chains for logical correctness and chart grounding.

Table~\ref{tab:annotation_stats} summarizes the scale of the annotation
process and the agreement statistics observed during the audit stage.

During the audit stage, disagreements typically arose from ambiguous
variable bindings, inconsistent axis references, or incomplete chart
context in the premise instantiation. These cases were resolved through
discussion between reviewers and the lead annotator, after which the
template definitions were updated to enforce stricter variable typing
and predicate constraints.

Overall, the high reviewer agreement and low template revision rate
suggest that the structured premise design and annotation guidelines
provided a consistent framework for decomposing chart reasoning tasks.
The full annotation rubric, validation scripts, and template library
will be released with the dataset.

\bibliographystyle{splncs04}
\bibliography{main}
\end{document}